\documentclass{article} % For LaTeX2e
\usepackage{iclr2027_conference,times}

\usepackage{amsmath,amsfonts,bm}

\def\eqref#1{equation~\ref{#1}}
\def\1{\bm{1}}

\DeclareMathAlphabet{\mathsfit}{\encodingdefault}{\sfdefault}{m}{sl}
\SetMathAlphabet{\mathsfit}{bold}{\encodingdefault}{\sfdefault}{bx}{n}

\usepackage{hyperref}
\usepackage{url}

\usepackage{graphicx} % 插入图片必需
\usepackage{caption}  % 允许在非浮动环境中生成标准图注
\usepackage{tabularx}
\usepackage{adjustbox}
\usepackage{subcaption}
\usepackage{amsmath}
\usepackage{hyperref}
\usepackage{comment}
\usepackage{cleveref}
\usepackage{xspace}
\usepackage{float}
\usepackage{bbding}
\usepackage{multirow}
\usepackage{xcolor} % 用于颜色
\usepackage{pifont}
 
\usepackage{amssymb}
\usepackage{soul}
\usepackage[normalem]{ulem} % [ZZ-1] ADDED: 提供 \sout，用于删除线批注；normalem 避免改写 \emph
\usepackage{graphicx}
\usepackage{booktabs}
\usepackage{array} % needed for p{...}<{\centering} author columns
\usepackage{enumitem}
\usepackage{tcolorbox}
\tcbuselibrary{breakable}
\definecolor{stmmdoorbell}{HTML}{1F77B4}
\definecolor{stmmalarm}{HTML}{D62728}
\newcommand{\emphline}[1]{\textbf{#1}}
\def\onedot{.\xspace}
\def\eg{\emph{e.g}\onedot} 
\def\ie{\emph{i.e}\onedot}

\def\etal{\emph{et al}\onedot}
\definecolor{nvidiagreen}{RGB}{119,185,0}
\definecolor{pgonegreen}{RGB}{17,166,80}
\definecolor{stmmalarm}{HTML}{2CA02C}    % 学术质感深绿色
\definecolor{stmmdoorbell}{HTML}{D4AC0D} % 学术质感琥珀黄，白底清晰可见！
\definecolor{AudioMaxColor}{HTML}{4E81A9}
\definecolor{AudioMinColor}{HTML}{D3A53E}
\definecolor{TrainColor}{HTML}{4A90E2}
\definecolor{TestColor}{HTML}{7F8C8D}
\definecolor{ThemeOrange}{HTML}{FCA64E}
\definecolor{ThemeBlue}{HTML}{41B7FB}
\newlength\savewidth\newcommand\shline{\noalign{\global\savewidth\arrayrulewidth
  \global\arrayrulewidth 1pt}\hline\noalign{\global\arrayrulewidth\savewidth}}
\definecolor{forestgreen}{RGB}{34, 139, 34}

\newcommand\todohao[1]{\colorbox{yellow}{#1}}
\newcommand{\reffig}[1]{Figure~\ref{#1}}
\newcommand{\reftab}[1]{Table~\ref{#1}}
\newcommand{\refsection}[1]{Section.~\ref{#1}}
\newcommand{\Rnum}[1]{\uppercase\expandafter{\romannumeral #1\relax}}
\newcommand{\textitbf}[1]{\textbf{\textit{#1}}}
\title{From Instruction to Event: Sound-Triggered Mobile Manipulation}

\author{\texorpdfstring{%
    \mdseries % cancel ICLR sty's \bf on \@author
    \begin{tabular}{@{}c@{}}
      \begin{tabular}{@{} p{0.22\textwidth}<{\centering} p{0.22\textwidth}<{\centering} p{0.22\textwidth}<{\centering} p{0.22\textwidth}<{\centering} @{}}
        {\large Hao Ju} & {\large Shaofei Huang} & {\large Hongyu Li} & {\large Zihan Ding} \\
        {\small University of Macau} & {\small Hefei University of Technology} & {\small Beihang University} & {\small Beihang University}
      \end{tabular} \\[3ex]
      \begin{tabular}{@{} p{0.28\textwidth}<{\centering} p{0.28\textwidth}<{\centering} p{0.28\textwidth}<{\centering} @{}}
        {\large Si Liu} & {\large Meng Wang} & {\large Zhedong Zheng} \\
        {\small Beihang University} & {\small Hefei University of Technology} & {\small University of Macau}
      \end{tabular}
    \end{tabular}%
}{Hao Ju, Shaofei Huang, Hongyu Li, Zihan Ding, Si Liu, Meng Wang, Zhedong Zheng}}

\usepackage[normalem]{ulem}                        % [ZZ] strikethrough for \zzdel/\zzrep

\iclrfinalcopy % show real authors + remove line numbers (camera-ready mode)
\begin{document}

\maketitle
% Keep under-review header; do NOT use "Published as a conference paper at ICLR 2027"
\lhead{Under review}
\begin{center}
  \vspace{-0.42in}
  \includegraphics[width=\textwidth]{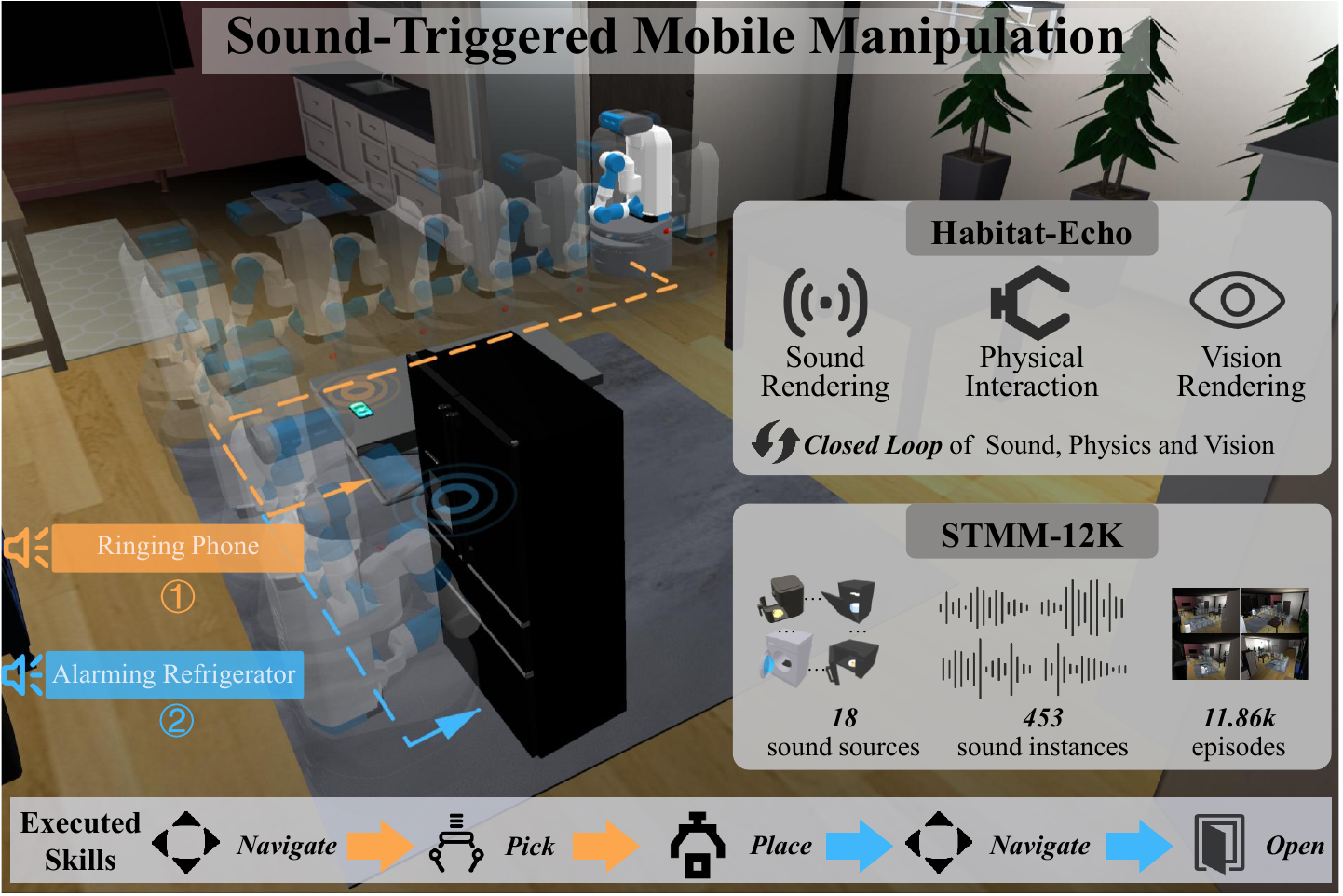}
  \vspace{-0.25in} % 微调图片与图注的间距
  \captionof{figure}{
  We introduce the problem of \textbf{Sound-Triggered Mobile Manipulation (STMM)}, where robots are required to infer task goals from environmental sound events, ground the corresponding sound-emitting targets, and execute appropriate physical interactions without explicit per-episode instructions. In multi-event scenes, robots must further prioritize and sequentially resolve concurrent events. To support this new task, we develop \textbf{Habitat-Echo}, a simulation platform that unifies sound rendering, visual perception, and physical interaction in a closed loop, and establish \textbf{STMM-12K}, a benchmark containing 18 sound-source categories, 453 sound instances, and 11.86k single-event and multi-event episodes.
  }
  \label{fig:motivation}
\end{center}
% \vspace{0.1in} % 微调图注与摘要的间距

\begin{abstract}
  Current mobile manipulation research predominantly follows an instruction-driven paradigm, where robots rely on predefined textual commands to execute tasks. 
  However, this setting confines robots to a passive role, limiting robotic autonomy and the ability to react to dynamic environmental events.
  To address these limitations, we introduce Sound-Triggered Mobile Manipulation (STMM), where robots must actively perceive and interact with sound-emitting objects without explicit action instructions.
  To support STMM, we develop Habitat-Echo, a simulation platform that integrates sound rendering with physical interaction. 
  We further propose a hierarchical baseline that translates high-level planning into low-level executions, where a task planner predicts a skill chain for policy models to execute sequentially.
  Experiments indicate the feasibility of perceiving auditory events and executing corresponding physical interactions without explicit instructions. 
  Notably, in challenging multi-event scenarios, the robot successfully isolates the primary sources from overlapping acoustic interference to execute the first interactions, and subsequently proceeds to manipulate the secondary objects.
  These new challenges of planning-to-execution position STMM as a measurable research direction.
  \textit{The code and datasets will be released upon acceptance.}
\end{abstract}

\section{Introduction}
% background of (mobile) manipulation
% 现有移动操作（mobile manipulation）已在长时序任务执行方面取得了显著进展，使智能体能够在复杂环境中完成导航与物体交互。然而，大多数现有方法均基于 instruction-driven paradigm，即任务目标由用户通过显式文本指令提前给定，智能体在接收指令后执行相应操作。这一范式适用于用户需求已被明确表达的场景，例如指定物体搬运或预定义任务执行。然而，其隐含前提是：任务目标在执行前已经被完整描述。
% 在真实日常环境中，大量任务并非源于用户事先给出的指令，而是由环境中的突发事件触发。例如电话响起、门铃触发或水流声持续等，这些事件往往具有突发性和不确定性，用户难以及时提供明确指令。这就要求智能体具备在事件发生后自主感知并作出响应的能力，即 event-driven mobile manipulation。然而，现有以指令为中心的移动操作方法难以直接适用于此类场景。
% [ZZ-11] PARAGRAPH MERGE: 原 intro 第 1 段与第 2 段合并（"Despite these advances" 直接回指第 1 段，
%         断开反而是逻辑跳跃）。全 intro 由 6 段收到 4 段。
Recent advances in embodied AI have led to remarkable progress in mobile manipulation, enabling robots to navigate in unstructured environments and physically interact with surrounding objects~\citep{song_survey_arxiv25,bai_towards_arxiv25,yu_far_arxiv25,chen2026abot}.
Such synergy between mobility and interaction empowers robots to perform long-horizon household tasks such as object rearrangement, multi-stage interaction, and sequential task execution~\citep{ji_robobrain_cvpr25,huang_roboground_cvpr25,li_driverse_mm25,adri_multi_icml25,song_towards_cvpr25,lin_vlnverse_arxiv25}, pushing embodied robots closer to general-purpose household assistance.
Despite these advances, most existing mobile manipulation works are built upon an instruction-driven paradigm, where the task goal is explicitly specified by the user through instructions~\citep{wu_momanipvla_cvpr25,shah_bumble_icra25,yan_dynamic_ral25,zhang_moma_arxiv25,liu_vln_mm25,li_ali_mm25}, and multimodal contextual cues~\citep{wang_roboomni_iclr26}. 
While this paradigm excels at completing user-specified chores, it also implicitly assumes that the task objective is known before execution begins.
However, in real-world household environments, many tasks are triggered by environmental events that require a timely response. 
For example, a ringing refrigerator may prompt robots to approach and open it, and a ringing phone may indicate that the device needs to be turned off (see~\reffig{fig:motivation}).
The unpredictable nature of these events makes it difficult for a user to monitor the scene and issue instructions in time.
This limitation highlights the urgent need for \textit{event-driven mobile manipulation}, where an intelligent robot should be able to proactively perceive and respond to environmental events.

% 在多种环境信号中，声音是触发事件的重要载体。在实际生活中，许多关键事件往往首先通过声音暴露，例如铃声、碰撞声或水流声等。与视觉信号相比，声音具备更强的全局可感知性，即使目标不在视野范围内或处于低光照条件下，智能体仍可通过音频信号感知事件的发生。同时，声音不仅提供事件发生的时间信息，还包含一定的语义提示和空间线索，有助于智能体推断事件类型并定位潜在目标。
% 基于此，我们关注一类由声音触发的事件响应问题，并提出 sound-triggered mobile manipulation。在该设定中，智能体不再依赖逐任务的显式文本指令，而需要基于听到的声音事件以及场景上下文，自主识别相关目标并执行相应交互。
% [ZZ-13] PARAGRAPH MERGE: 原第 3 段（声音为何适合做触发信号）与原第 4 段（任务分类）合并，
%         因为 "Concretely, we categorize..." 是对 "we introduce STMM" 的直接展开。
Among various environmental signals, sound serves as a natural trigger for event-driven interaction.
% [ZZ-14] 原句是比较结构悬空（"compared to visual cues" 挂在了 illumination 上），改写。
% \zzrep{Sound is omnidirectional and independent of illumination compared to visual cues.}{Unlike visual cues, sound is omnidirectional and independent of illumination.}
Unlike visual cues, sound is omnidirectional and independent of illumination.
In fact, many of these events can be perceived acoustically long before they enter the visual field. % [ZZ-15] 删填充词
Robots can therefore detect critical events even when the sounding target lies outside its field of view or under poor lighting conditions. 
Moreover, sound signals not only indicate the occurrence of an event, but also provide partial \textit{semantic} cues about the event type and \textit{spatial} cues about the location of the sound source. 
Motivated by this, we introduce the problem of \textbf{Sound-Triggered Mobile Manipulation (STMM)}, where robots must autonomously ground the sound source and execute appropriate interactions based on the sound event and scene context. % [ZZ-17] 全文 "novel" 出现 4 次，此处删一个
Unlike conventional instruction-driven mobile manipulation, STMM requires robots to infer the task goal from environmental events rather than receive instructions or context directly from a human user. % [ZZ-18] DELETE "(see~\reffig{fig:motivation})"：Fig.1 已在上文引用，重复指向
Concretely, we categorize sound-triggered mobile manipulation into two fundamental interaction types:
(1) \textbf{object relocation}, where the robot localizes a sounding rigid object (\eg, a phone) and alters its spatial position via pick-and-place skills, and (2) \textbf{state transition}, where the robot localizes a sounding functional or articulated object (\eg, a refrigerator) and manipulates it to transition its internal state, \eg, opening or closing.
Building upon this, we investigate both single-event and more challenging multi-event settings.
In the multi-event setting where multiple sound events occur simultaneously, robots must recognize concurrent events, determine the corresponding interactions, and compose them into an ordered execution plan according to a specified event priority.
% [ZZ-Q, 重要] "specified event priority" 是谁指定的、以什么形式进入 agent 的输入？
%   如果 priority 是一个显式符号输入，那么 STMM "不依赖显式指令" 的核心主张在多事件设定下不成立。
%   请在 Method 里明确写出 t=0 时 agent 究竟收到哪些量（含 priority 的编码方式），并在 Problem Formulation 里承认这一点。
Overall, sound-triggered mobile manipulation requires robots to jointly reason about \textit{what} event is happening, \textit{where} the relevant sounding target is located, and \textit{how} the correct interaction should be carried out, while additionally handling task scheduling in multi-event scenarios, making the problem particularly challenging.

% 然而，现有具身仿真平台通常仅支持单一模态能力：要么关注视觉与物理交互（如移动操作模拟器），要么关注音频感知（如声源定位环境），缺乏一个能够同时支持 空间音频感知与物理交互 的统一平台。这一缺失在很大程度上限制了 sound-triggered mobile manipulation 的系统性研究。
% 为此，我们构建了 Habitat-Echo，一个面向声音触发移动操作的音视物理一体化仿真平台。该平台在现有具身仿真环境的基础上，引入空间音频建模能力，并扩展了发声物体与可交互对象，使智能体能够在同一环境中完成声音感知、目标定位以及物理操作的闭环交互。
However, existing embodied simulators either focus strictly on visual-physical interaction (\eg, mobile manipulation simulator~\citep{xavier_habitat_iclr24}) or visual-sound sensing (\eg, sound navigation environment~\citep{chen_soundspaces_eccv20}).
There is a lack of an integrated environment where robots can perceive spatial sound and physically interact with sound-emitting targets within a closed loop, hindering systematic research on sound-triggered mobile manipulation.
To bridge this gap, we develop \textbf{Habitat-Echo}, a unified sound-visual-physical simulation platform.
By integrating spatial sound rendering into existing interactive embodied environments and extending them with abundant sound-emitting interactive assets, Habitat-Echo enables a closed-loop simulation of sound perception, sound source grounding, and physical interaction within a single unified platform.

% 基于 Habitat-Echo，我们进一步构建了首个面向 sound-triggered mobile manipulation 的基准数据集，涵盖多种发声物体与交互场景。在该基准上，我们设计了一种层级式 baseline 框架来自顶向下的将复杂的动态发声事件拆解为：高层模块作为任务规划器，基于音频与视觉输入推理任务目标并生成可执行的技能序列（skill chain）；低层模块由一组轻量级策略模型组成，分别负责执行具体的导航与操作技能。该 baseline 并非旨在完全解决该任务，而是作为初步探索，验证任务设定的可行性，并揭示在声音感知、目标定位与多阶段交互方面的关键挑战。
Built on Habitat-Echo, we further establish \textbf{STMM-12K}, the first benchmark for sound-triggered mobile manipulation, covering diverse sounding objects and interaction scenarios under both single-event and multi-event settings.
To provide a first-step solution to this new problem, we design a hierarchical baseline framework that decomposes sound-triggered mobile manipulation into high-level acoustic reasoning and low-level physical execution. 
For high-level acoustic reasoning, a task planner is leveraged to translate sound-visual observations into an executable skill chain and organize it according to the specified event priority. 
For low-level physical execution, a set of lightweight policy models is invoked sequentially to carry out each required skill according to the predicted skill chain. 
Each policy model focuses on a specific skill, such as navigation, pick, place, and open door, enabling specialized control for diverse interaction types. 
Experiments on STMM-12K shows: (1) completion success rates with ground-truth plans and privileged target positions reach 86.13\% for object relocation but only 30.38\% for state transition. (2) Supervised task planners achieve planning success rates up to 99.03\% on object relocation but 69.50\% on the multi-event setting.
Together, these results highlight the challenges, particularly the cascade completion for state transitions and the task planning for the multi-event setting. Our contributions are summarized as follows:
\begin{itemize}[leftmargin=0pt]
\item We propose Sound-Triggered Mobile Manipulation (STMM), a novel event-driven embodied task in which robots are required to respond to sound events physically without relying on explicit per-episode instructions. We define two interaction types, object relocation and state transition, and instantiate each under both single-event and multi-event settings.
\item We develop Habitat-Echo, a unified embodied simulation platform that unifies spatial sound rendering, visual observation rendering, and interactive physical interaction in a close loop. Based on Habitat-Echo, we establish STMM-12K, the first benchmark for STMM containing 11.86k episodes (10.47k train / 1.39k test), 453 sound instances, and 18 sound sources. 
\item  Experimental results on STMM-12K show the feasibility and challenges of our hierarchical baseline. Specifically, the baseline achieves a 71.05\% completion success rate on single-event object relocation, showing the feasibility of relocating rigid sound sources. Meanwhile, single-event state transitions pose completion challenges (28.67\%), while the multi-event setting presents additional task planning challenges (69.50\%).

% We further propose a hierarchical baseline framework that offers an initial solution and reveals the core challenges of this new embodied task. 
% Supervised planning reaches 99.0\% accuracy, yet completion is 71.1\% for object relocation and 28.7\% for state transition, and only 7.75\% under dual-source interference.

\end{itemize}

\begin{figure*}[!t]
    % \vspace{-0.1in}
    \centering
    \includegraphics[width=\linewidth]{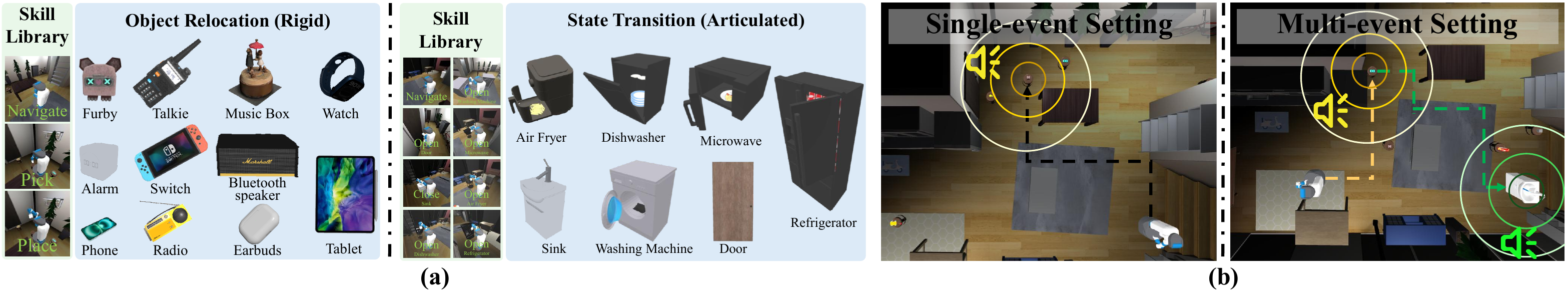}% 中括号中的为调节图片大小
    \vspace{-.15in}
    \caption{
    {
    Task taxonomy and event settings of STMM.
    (a) STMM comprises two interaction types according to the physical properties of sound-emitting targets: \textit{object relocation} for rigid sources and \textit{state transition} for articulated sources.
    (b) Both interaction types are instantiated under \textit{single-event} and \textit{multi-event} settings, and the latter requires robots to resolve concurrent events sequentially according to a specified priority.}
    % {Different settings of Sound-Triggered Mobile Manipulation (STMM).} 
    % {(a)} {Object Relocation in the single-event setting} requires robots to interact with the sound source (a rigid object) via Navigate, Pick, and Place skills.
    % {(b)} {State Transition in the single-event setting} requires robots to interact with the sound source (an articulated object) via Navigate, Open Door, and Close Sink skills.
    % {(c)} {Multi-event Setting} requires robots to interact with the sound sources sequentially.
}
    \label{fig:task}
    \vspace{-.15in}
\end{figure*}

\section{Related Work}
% \noindent\textbf{Related Work: Sound in Embodied AI.} 
Sound provides crucial spatial and contextual information for embodied agents. Existing embodied AI leverages sound primarily for either navigation or manipulation. 
Below, we review these two lines of work in turn.
In Sound-Visual Navigation~\citep{chen_soundspaces_eccv20}, methods address sound rendering (\eg, semantic emissions~\citep{chen_semantic_cvpr21}, moving sources~\citep{younes_catch_ral23}, distractors~\citep{yu_sound_arxiv22}) or cross-modal learning~\citep{wang_towards_nips22,kondoh_embodied_cvpr25,wang2022align}, yet largely neglect physical interaction.
% 
% In robotic manipulation, sound typically serves as an informative modality to guide physical processes.
% \citep{wang_sound_corl25} utilizes MMAudio~\citep{ruan_mm_cvpr23} to generate the sound of pouring liquid into a bottle to assist the process. RoboOmni~\citep{wang_roboomni_iclr26} extends textual instructions into audio contexts and uses these contexts to infer human intent, enabling interactive communication between human and robot.
% Additionally, \citep{guo_omnivla_icra26} employs beamforming to convert acoustic signals into 2D masks, complementing visual observations. 
% Unlike prior omni-modal manipulation methods that use sound or speech as contextual cues for inferring user intent, STMM treats the spatial acoustic event itself as the task trigger and the sound-emitting entity as the physical interaction target, requiring navigation, source grounding, and manipulation in a closed loop.
Works in robotic manipulation utilize sound merely as auxiliary context, such as for process monitoring~\citep{wang_sound_corl25}, intent inference~\citep{wang_roboomni_iclr26}, or 2D masking~\citep{guo_omnivla_icra26}. Unlike these approaches that treat sound as a complementary cue, STMM casts the sound event as the explicit task trigger and the sound source as the physical target, unifying grounding, navigation, and manipulation into a closed loop.
\textit{We defer further discussion to the supplementary material.}

\section{Sound-Triggered Mobile Manipulation}
\label{sec:STMM}

\subsection{Problem Formulation}
% Unlike conventional instruction-driven mobile manipulation, which is typically formulated as learning a policy $\pi(a_{t} \mid O_t, L)$ conditioned on an explicit instruction $L$, Sound-Triggered Mobile Manipulation (STMM) removes the explicit instruction that specifies the task goal.
% Instead, we formulate STMM as a partially observable Markov decision process.
% % [ZZ-27] 声明用 POMDP 却没给元组和奖励，审稿人必问。这里补形式化并留 TODO。
% \zzadd{$\mathcal{M}=\langle \mathcal{S}, \mathcal{A}, \mathcal{T}, \mathcal{R}, \Omega, \mathcal{O}, \gamma \rangle$}
% % [ZZ-TODO] 请补齐每个分量的定义，尤其是：\mathcal{S} 是否含物体位姿？\mathcal{R} 的 event-resolution
% %   奖励项是什么？终止条件如何与技能成功检测绑定？否则 "POMDP formulation" 只是标签。
% At each time step $t$, the robot only receives a multimodal observation $O_t=\{V_t,S_t,P_t\}$, where $V_t$, $S_t$, and $P_t$ denote visual observations, sound observations, and proprioceptive states, respectively.
% In response to environmental sound events $E_{sound}$, the robot must implicitly infer the task goal $\mathcal{G}$ from its observations and execute actions $a_t \in \mathcal{A}$ over time to interact with the relevant target.
% The episode is considered successful if the event-conditioned interaction induced by $E_{\text{sound}}$ is accomplished (see ~\refsection{sec:skill}).
% Concretely, as shown in~\reffig{fig:task}, depending on the physical properties of the sound sources associated with $E_{sound}$, we categorize the required interactions to fulfill $\mathcal{G}$ into two fundamental types:
Unlike conventional instruction-driven mobile manipulation, Sound-Triggered Mobile Manipulation (STMM) requires the robot to respond to sound events without a complete per-episode task instruction.
We model STMM as a partially observable Markov decision process,
$\mathcal{M}=\langle \mathcal{S},\mathcal{A},\mathcal{P},
\mathcal{R},\Omega,\gamma\rangle$,
where $\mathcal{S}$, $\mathcal{A}$, and $\Omega$ denote the state,
action, and observation spaces, respectively.
$\mathcal{P}$ specifies the joint distribution of the next state
and observation given the current state and action,
while $\mathcal{R}$ and $\gamma$ denote the reward function
and discount factor, respectively.
At each time step $t$, the robot receives a multimodal observation
$O_t=\{V_t,S_t,P_t\}\in\Omega$, where $V_t$, $S_t$, and $P_t$
denote visual observations, sound observations, and proprioceptive
states, respectively.
From these observations, the robot must predict the task goal $\mathcal{G}$ associated with sound events $E_{\text{sound}}$ and execute actions $a_t\in\mathcal{A}$ to navigate and interact with the relevant sound sources.
In the multi-event setting, event priorities are additionally provided (see~\refsection{sec:task}).
Under our evaluation protocol, an episode is successful only if the required skill chain is correctly predicted and all constituent skills are successfully executed in order (see~\refsection{sec:skill}).
As shown in~\reffig{fig:task}, we categorize the interactions into two fundamental types according to the physical properties of the corresponding sound sources:

\noindent\textbf{Object Relocation.} 
% Here, the sounding target is a rigid object (\eg, a ringing phone in~\reffig{fig:task}a), and the goal is to resolve the event by changing its spatial position (\eg, answering the ringing phone).
% To accomplish this, the agent must localize and approach the relevant sound source, and then perform object-centered manipulation such as picking and placing.
% The main challenge lies in translating cross-modal cues to executable object interaction.
% 
For rigid sound sources (\eg, a ringing phone in~\reffig{fig:task}a), the robot resolves the event by relocating the object. This requires localizing, approaching, and manipulating the source, which requires translating cross-modal observations into executable interactions.

\noindent\textbf{State Transition.} 
% Here, the sounding target is a functional or articulated object governed by object-specific kinematic constraints (\eg, a running faucet in~\reffig{fig:task}b), and the goal is to resolve the event by changing its internal physical state (\eg, turning off the faucet).
% Compared to object relocation, since the agent must operate along the object's functional axis or articulated trajectory, this interaction type imposes stronger constraints on local geometry and manipulation precision.
% For example, the agent must precisely grasp a faucet handle and rotate it along its functional axis to stop the running water.
% 
For articulated sound sources (\eg, a beeping washing machine in~\reffig{fig:task}b), the robot resolves the event by transitioning the object's internal physical state. Unlike spatial relocation, this interaction requires operating along specific kinematic trajectories, imposing stricter constraints on local geometry and manipulation precision. For example, opening a washing machine necessitates precisely grasping and rotating its handle along the functional axis.

\subsection{Task Setting}
\label{sec:task}
Beyond the interaction types defined above, we further design single-event and multi-event settings for STMM, which differ in the complexity of source localization and task execution.

\noindent\textbf{Single-event Setting.}
In the single-event setting, only one sounding event is present in the scene. 
The robot is required to infer the event-conditioned goal from the multimodal observation, localize the relevant target, and execute the corresponding interaction to resolve the event. 
Depending on the physical properties of the sounding target, the required interaction can take the form of either object relocation or state transition.
% [ZZ-Q] 单事件设定里只有一个声源，且声源类别到技能链基本是 1:1 映射（如 phone -> Navigate-Pick-Place）。
%   那么 "规划" 在单事件下退化成声源类别分类。审稿人会据此质疑 planning SR 98.78% 的含金量。
%   建议在正文里主动说明这一退化，并补一个 "同类别多实例 / 同实例多可能动作" 的更难子集。

\noindent\textbf{Multi-event Setting.}
In the multi-event setting, multiple sound events occur simultaneously, resulting in acoustic interference and sequential decision-making challenges. 
Given an event-priority specification, the robot must identify the concurrent events, associate each event with the corresponding interaction, and execute the resulting skill chains in the prescribed order.
After the first event is resolved, its corresponding sound disappears and the acoustic scene changes, requiring the robot to continue interacting with the remaining target under an updated environment state.

\subsection{Skill Space}
\label{sec:skill}
Directly mapping raw observations to low-level actions entangles high-level reasoning with physical execution, increasing learning burden and hindering proactive decision-making. To address this, STMM decouples the entangled process into reusable skill primitives, each serving as an atomic sensorimotor mapping for specific interactions.
Specifically, the skill space is defined as:
\begin{equation}
\mathcal{K}=\{\texttt{Navigate},\texttt{Pick},\texttt{Place},\texttt{CloseSink},...,\texttt{OpenRefrigerator}\}.
\end{equation}
Here, \texttt{Navigate} denotes approaching the relevant sounding target.
\texttt{Pick} and \texttt{Place} correspond to rigid-object relocation via the robot gripper. 
\texttt{CloseSink} and \texttt{Open Refrigerator} correspond to state-transition interactions. 
The detailed execution spaces and termination criteria are elaborated in the supplementary materials.
Under this abstraction, different events can be solved through different skill compositions, \ie, the \textit{Skill Chain}.
An event is considered resolved successfully if and only if all constituent skills are successfully executed in sequence.
Object relocation is typically instantiated as:
\begin{equation}
[\texttt{Navigate}\rightarrow \texttt{Pick} \rightarrow \texttt{Place}],
\end{equation}
whereas state transition is instantiated as:
% [ZZ-32] "[\texttt{Navigate}\rightarrow \texttt{CloseSink}]~,~...., ~[...]" 的 ",....," 排版很糟
\begin{equation}
[\texttt{Navigate}\rightarrow \texttt{CloseSink}]~,~...., ~[\texttt{Navigate}\rightarrow \texttt{OpenRefrigerator}].
\end{equation}
For multi-event scenes, the complete task is further represented as the ordered composition of multiple skill chains conditioned on a specified event priority. 
In this way, the skill chain serves as a structured bridge from latent event interpretation to executable embodied responses.

\section{Platform and Benchmark}
\subsection{Habitat-Echo Simulation Platform}
\begin{figure*}[t]
    \centering
    % \vspace{-.1in}
    % Make subfigure labels closer to the contents
    \captionsetup[subfigure]{skip=1pt}

    % ============================================================
    % Left column: (a) Simulator comparison + (b) YCB objects
    % ============================================================
    \begin{minipage}[c]{0.40\textwidth}
        \centering

        % --------------------------------------------------------
        % (a) Simulator comparison
        % --------------------------------------------------------
        \begin{subfigure}[c]{\linewidth}
            \centering

            \resizebox{0.98\linewidth}{!}{%
                \begin{tabular}{ccc}
                    \shline

                    Simulator
                    & Sound Rendering
                    & Physical Interaction \\

                    \hline

                    Habitat~\citep{andrew_habitat_nips21}
                    & \textcolor{red}{\ding{55}}
                    & \textcolor{nvidiagreen}{\ding{51}} \\

                    AI2-THOR~\citep{kolve_ai2thor_arxiv17}
                    & \textcolor{red}{\ding{55}}
                    & \textcolor{nvidiagreen}{\ding{51}} \\

                    Isaac~\citep{makoviychuk_isaac_arxiv21}
                    & \textcolor{red}{\ding{55}}
                    & \textcolor{nvidiagreen}{\ding{51}} \\

                    SoundSpaces~\citep{chen_soundspaces_eccv20}
                    & \textcolor{nvidiagreen}{\ding{51}}
                    & \textcolor{red}{\ding{55}} \\

                    \hline

                    Habitat-Echo (Ours)
                    & \textcolor{nvidiagreen}{\ding{51}}
                    & \textcolor{nvidiagreen}{\ding{51}} \\

                    \shline
                \end{tabular}%
            }

            \caption{}
            \label{fig:simulator}
        \end{subfigure}

        % Space between (a) and (b)
        \vspace{0.005in}

        % --------------------------------------------------------
        % (b) YCB object dataset
        % --------------------------------------------------------
        \begin{subfigure}[c]{\linewidth}
            \centering

            % Same width as the table above
            \includegraphics[
                width=0.98\linewidth,
                keepaspectratio
            ]{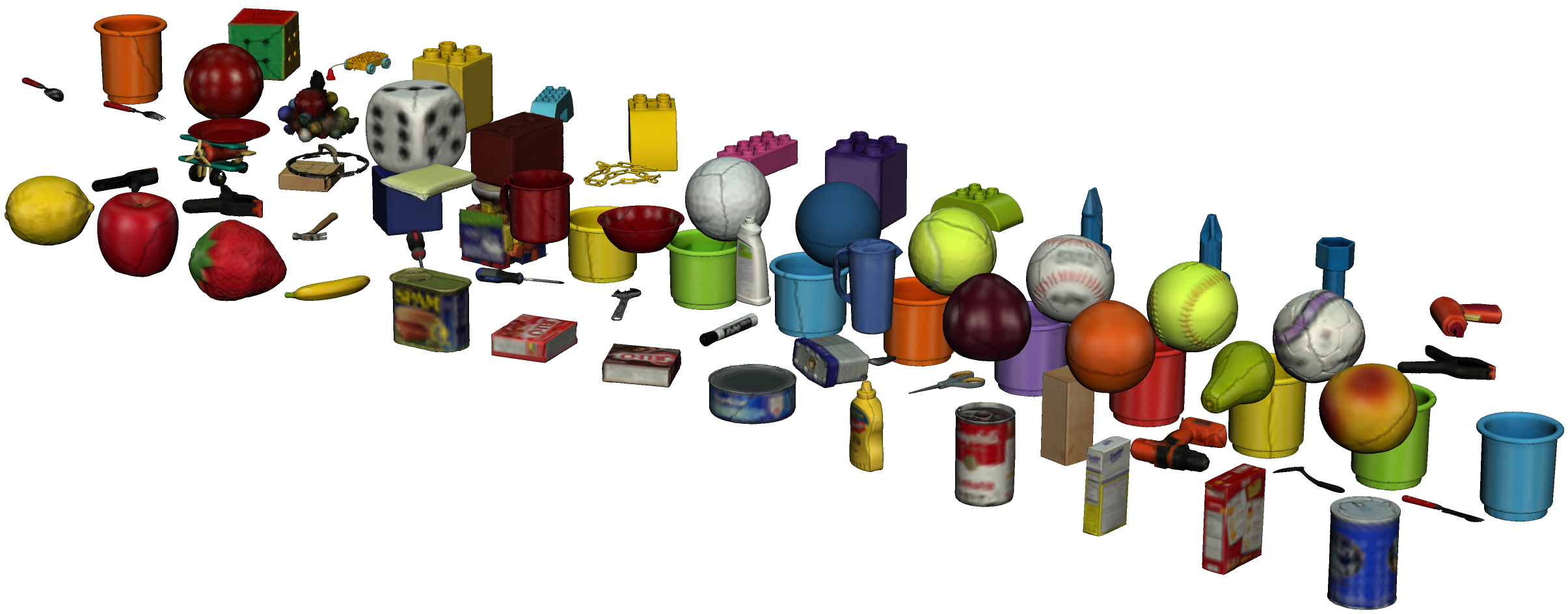}

            \caption{}
            \label{fig:ycb_object_dataset}
        \end{subfigure}

    \end{minipage}
    \hfill
    % ============================================================
    % (c) Audio asset distribution
    % ============================================================
    \begin{subfigure}[c]{0.285\textwidth}
        \centering

        \includegraphics[
            width=0.96\linewidth,
            keepaspectratio
        ]{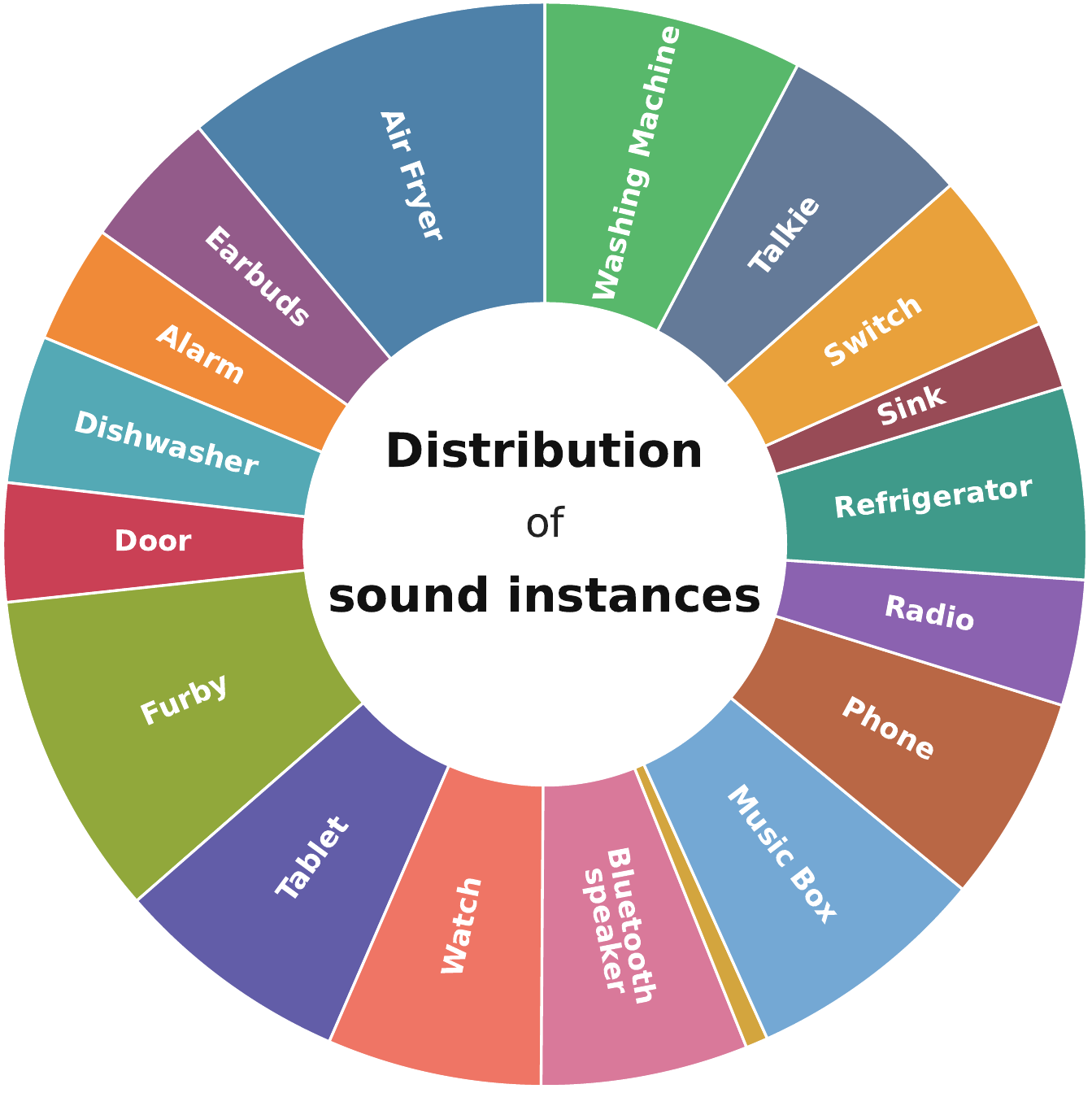}

        \caption{}
        \label{fig:audio_asset}
    \end{subfigure}
    \hfill
    % ============================================================
    % (d) Episode statistics
    % ============================================================
    \begin{subfigure}[c]{0.285\textwidth}
        \centering

        \includegraphics[
            width=0.96\linewidth,
            keepaspectratio
        ]{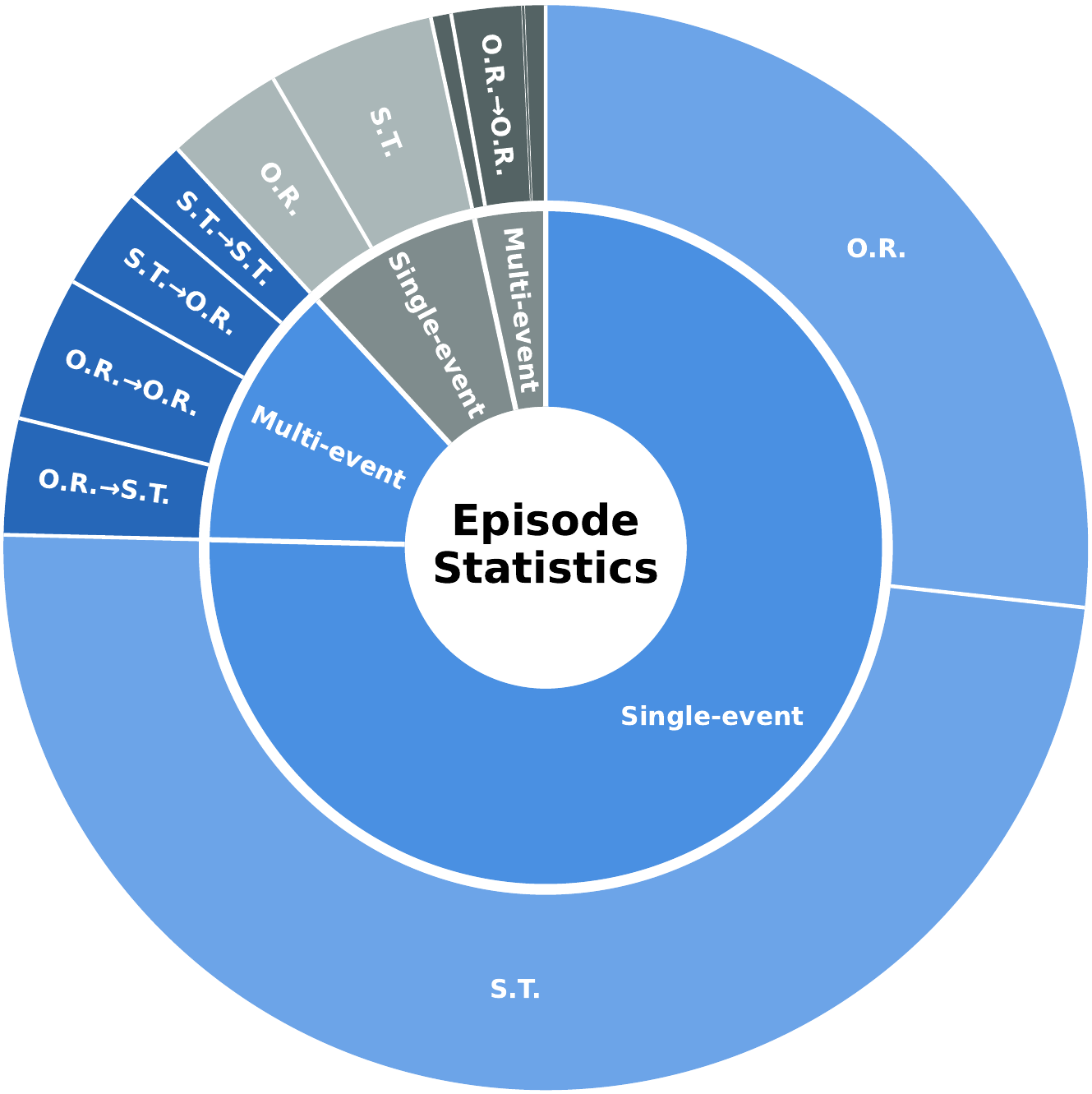}

        \caption{}
        \label{fig:dataset}
    \end{subfigure}

    % ============================================================
    % Main caption
    % ============================================================
    \vspace{-0.1in}

    \caption{
        {(a) Comparison of different simulators.}
        Habitat-Echo supports both sound rendering and physical interaction.
        {(b) Visual distractor objects.}
        STMM-12k includes \textit{77} additional visual distractor objects.
        {(c) Distribution of sound instances.}
        Instances are color-coded by category.
        {(d) Episode statistics.}
        \textcolor{TrainColor}{Train} and \textcolor{TestColor}{test} splits. (\textcolor{TrainColor}{10.47k} and \textcolor{TestColor}{1.39k} episodes)
        O.R. denotes Object Relocation, and S.T. denotes State Transition.
        Interaction type of the first sound source $\rightarrow$ the second one.
    }
    \label{fig:habitat_echo_overview}

    \vspace{-.15in}
\end{figure*}

Existing simulators either prioritize visual-physical manipulation 
~\citep{kolve_ai2thor_arxiv17,makoviychuk_isaac_arxiv21,andrew_habitat_nips21} or strictly focus on sound-visual navigation~\citep{chen_soundspaces_eccv20,chen_soundspaces_nips22}, leaving sound perception and physical interaction disconnected (see~\reffig{fig:simulator}).
To support STMM, we develop Habitat-Echo, which couples spatial sound rendering with interactive object dynamics under a unified simulation state, further enabling sound-visual-physical closed-loop interaction.

\noindent \textbf{Sound-Physical Simulation.}
{
Habitat-Echo builds upon interactive indoor 3D environments~\citep{andrew_habitat_nips21}, {retaining visual rendering and physical interactions} while integrating the spatial sound capabilities of SoundSpaces~\citep{chen_soundspaces_eccv20}.
At each time step, the simulator retrieves the Room Impulse Response (RIR) corresponding to the current source–receiver {spatial} configuration and computes binaural sound observations from the source waveform.}
As the robot moves, the resulting sound observation changes consistently with the evolving spatial relationship between the robot and the sound source. 
Habitat-Echo further supports simultaneous rendering of multiple active sources, allowing overlapping acoustic events to be perceived within the same scene. 
Importantly, acoustic events are coupled with the physical states of their corresponding objects. 
Once an event is physically resolved, its associated sound is removed from the subsequent observation, producing an updated acoustic field for subsequent interaction. 
This coupling allows robot actions to affect not only the physical environment but also future sensory observations, forming a sound–visual–physical closed loop.

\noindent \textbf{Interactive Sound-Event Assets.}
To instantiate such closed-loop interactions, we extend the environment with a collection of sound-emitting interactive assets for STMM.
These assets cover both interaction types in \refsection{sec:STMM}, including graspable sounding rigid bodies (\eg, phones, alarms) and articulated objects (\eg, doors, sinks with faucet handles), providing an asset foundation for both object relocation and state transition tasks.
% To enhance physical realism, these assets include coupled dynamic elements; for instance, the sink features water flow synchronized with the rotation of the faucet handle and refrigerators are filled with soda cans.
To enhance physical realism, these assets include coupled dynamics (\eg, water flow syncing with faucet handles) and populated interiors (\eg, stocked refrigerators).
These assets are explicitly designed to act as sound sources and manipulation interaction targets, thereby closing the loop between perception and action.
We further populate scenes with silent instances from sound-source categories and unrelated visual distractors (see~\reffig{fig:ycb_object_dataset}), preventing agents from resolving events solely through object-category priors.

\noindent \textbf{Skill Interface.}
% Supporting the skill space $\mathcal{K}$ in ~\refsection{sec:skill}, Habitat-Echo provides standardized execution interfaces and underlying observation/action space support for each skill primitive (\eg, supporting \texttt{Navigate} termination based on sound-visual features, or \texttt{OpenDoor} operations bounded by joint constraints). 
% The platform allows for the independent initialization, termination, and success detection of each skill, thereby enabling the sequential scheduling and closed-loop evaluation of entire skill chains.
% 
Habitat-Echo exposes standardized interfaces for the skill space $\mathcal{K}$ defined in~\refsection{sec:skill}, including navigation, object relocation, and articulated-object manipulation.
Different skills operate over their corresponding observation and action spaces, such as audio-visual observations for \texttt{Navigate} and joint-constrained actions for \texttt{OpenRefrigerator}. 
Each skill is equipped with consistent initialization, termination, and success criteria, allowing heterogeneous low-level policies to be composed into executable skill chains.

\subsection{STMM-12K Benchmark}

STMM-12K is designed to evaluate whether an embodied agent can predict interaction goals from acoustic events and physically resolve events under diverse objects, sounds, and compositions.
The benchmark follows three complementary design principles: 
(1) \textit{interaction diversity}, covering both object relocation and state transition;
(2) \textit{acoustic generalization}, requiring robots to handle unheard instances of known sound events; 
(3) \textit{event complexity}, ranging from single-event episodes with {silent} distractors to multi-event episodes with {concurrent} sounds and sequential interactions.

\noindent\textbf{Dataset Construction.} 
STMM-12K covers the two interaction types defined in~\refsection{sec:STMM}, \ie, object relocation and state transition, under both single-event and multi-event settings. 
As shown in \reffig{fig:task}, the benchmark contains 18 interactive sound sources in total, including 11 graspable rigid objects such as phones and alarms, and 7 articulated objects such as doors, sinks, and household appliances. 
Each category is associated with multiple real-world sound recordings, yielding 453 acoustic instances and providing diverse realizations of the same semantic event.
To prevent agents from exploiting visual-category shortcuts, we introduce silent source-category distractors (see~\reffig{fig:task}) and unrelated visual distractors (see~\reffig{fig:ycb_object_dataset}). 
The former require agents to distinguish acoustically active targets from visually plausible alternatives, while the latter increase scene clutter, jointly encouraging acoustic grounding rather than category-level visual recognition.

% YCB objects~\citep{calli_ycb_icra15}
% Beyond visual distractors (see~\reffig{fig:ycb_object_dataset}), our benchmark introduces diverse interactive sounding objects for different tasks (see~\reffig{fig:task}).
% For object relocation, STMM-12K includes 11 graspable rigid objects, such as phones and alarms.
% For state transitions, we incorporate articulated objects, such as sinks and {refrigerators}, which feature kinematic joints constrained to fixed functional axes. To enhance physical realism, these objects include coupled dynamic elements; for instance, the sink features water flow synchronized with the faucet handle's rotation and refrigerators are filled with soda cans.
% For sound generalization, we collect a diverse library of real-world sounds, implementing a one-to-many mapping in which each sound-source category is associated with multiple sound instances (\todohao{453 instances in total}). We further partition these sound instances into strictly disjoint training and test sets, while keeping the semantic sound categories shared across splits.
% Therefore, sound recordings used during evaluation are never observed during training, requiring the robot to generalize to unseen acoustic instances rather than memorizing specific waveforms.
% [ZZ-TODO] 请补：声源类别数、每类声实例数、可交互物体数、场景数（当前只有一个散点式的 77 个视觉干扰物）。

\noindent\textbf{Episode Generation.} 
We employ a procedural randomization process to generate episodes for training and testing.
For each episode, a sound-source category is first sampled, followed by a specific recording instance from that category.
The robot and target objects (except for doors) are then placed at randomized valid locations, while infeasible configurations are rejected to ensure start-to-goal traversability and physical accessibility for the required manipulation.
In the multi-event setting, two sound sources are randomly placed and validated for physical accessibility and traversability. One is then randomly selected as the main source, and the priorities of sound sources are recorded.
% In each episode, the sound for the target object is determined by hierarchically sampling a sound category and then a specific sound instance. 
% Visual complexity is introduced by adding two types of distractors alongside the target object: silent instances of sound-source objects (see~\reffig{fig:task}) and purely visual distractors (see~\reffig{fig:ycb_object_dataset}).
% The robot and target objects (excluding doors) are randomly positioned. We validate episode solvability by ensuring start-to-goal traversability and frontal accessibility for manipulating interactive objects, particularly sound sources.
STMM-12K comprises 11.86k episodes in total, including 8.97k/0.99k train/test episodes for the single-event setting, and 1.50k/0.40k for the multi-event setting (see~\reffig{fig:dataset}).
The sound instances are strictly divided into training (311) and testing (142) sets.

\noindent\textbf{Evaluation Protocol.}
We evaluate STMM at both the task planning and task completion levels.
\textit{Planning Success Rate (PSR)} measures whether the agent correctly predicts the event-conditioned skill chain. 
\textit{Completion Success Rate (CSR)} adopts a stricter criterion that an episode is counted as successful only when the required plan is correct and every constituent skill is successfully executed in the prescribed order.

% Performance is evaluated via task planning and skill completion success rates, with episode success strictly requiring accurate planning followed by the sequentially correct execution of all prerequisite skills.
% [ZZ-39] 指标命名全文有 4 套说法：planning rate / planning accuracy / planning success rate / overall success rate + completion / episode success。建议在此处一次性定义 PSR / CSR 两个缩写，后文统一使用。
% We report two metrics: task planning success rate and task completion success rate, where completion success rate counts an episode as successful only if the predicted skill chain is correct and every constituent skill succeeds in order.
% [ZZ-Q] 请说明 CSR 是否以 PSR 成功为条件（table 中 CSR 恒 <= PSR，看起来是联合指标而非条件指标）。

\begin{figure*}[!t]%调节图片位置，h：浮动；t：顶部；b:底部；p:当前位置
    % \vspace{-.25in}
    \centering
    \includegraphics[width=.97\linewidth]{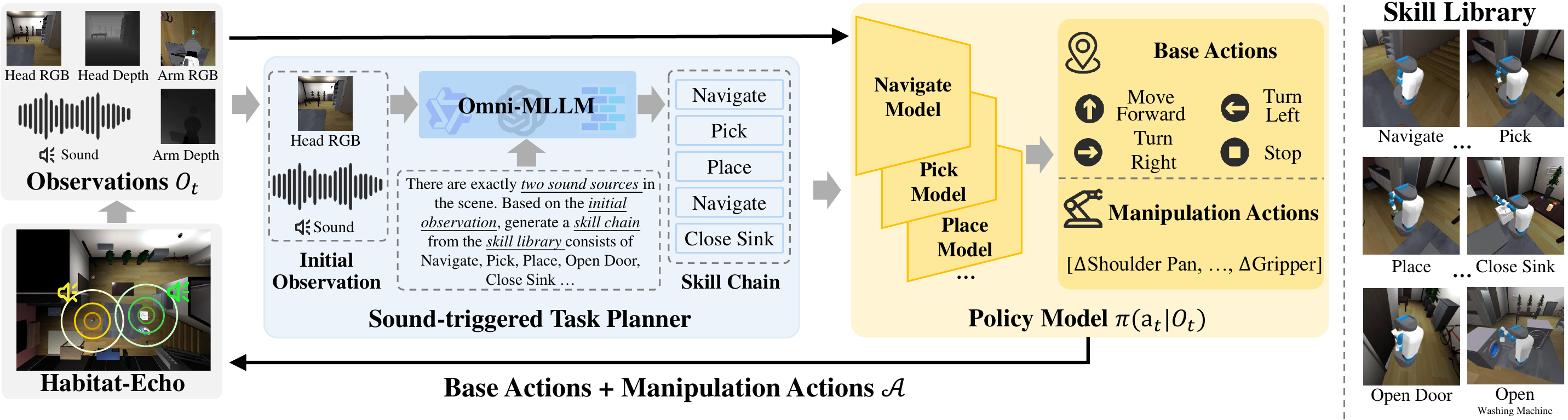}% 中括号中的为调节图片大小
    \vspace{-.15in}
    \caption{
    \emphline{Overview of the proposed baseline.} 
    The {sound-triggered task planner} processes the initial observation to reason and generate a high-level skill chain from the skill library \textit{(right)}. Guided by this chain, specialized policy models are sequentially activated to generate low-level actions and interact with Habitat-Echo.
    }
    \label{fig:baseline}
    \vspace{-.2in}
\end{figure*}
\section{Hierarchical Baseline for STMM}
The proposed baseline comprises two core components, \ie, a sound-triggered task planner and policy models (see~\reffig{fig:baseline}).
The sound-triggered task planner acts as the high-level reasoner, analyzing initial observations to predict a logical {skill chain}.
Following this chain, the policy models serve as the low-level executors, sequentially translating each assigned skill into concrete actions to complete the corresponding subtasks.

\subsection{Sound-triggered Task Planner}
We propose a sound-triggered task planner that operates as the high-level reasoning module. This module is responsible for translating environmental cues into a plan of skills, \ie, a skill chain (see~\reffig{fig:baseline}).
The input for the task planner is twofold, comprising initial sensory observations and a skill library.
Regarding observations, the planner utilizes the sound to reason about the target's category and relative orientation, while the head RGB provides an initial egocentric view to partially perceive the scene layout.
Crucially, to prevent information leakage, we restrict the planner's input to strictly align with the observation space of the Navigate skill, ensuring no additional information is accessed. 
Complementing these observations, the skill library is provided as a textual prompt or category embeddings for the task planner.
The library defines the semantic functionality of all available skills to support reasoning of the planner.
By combining these multi-modal observations with the skill descriptions, the task planner outputs an ordered chain of skills.
This chain acts as a high-level directive, guiding the sequential activation of the following policy models to complete the task.
% [ZZ-Q] "textual prompt or category embeddings" —— 两种输入形态对应两种不同的实现（Omni-MLLM 走 prompt，SSP 走 embedding）。请分别写明，否则读者无法判断 zero-shot 与 supervised 的比较是否同输入。

\subsection{Skill-Specific Policy Model}
Policy models serve as low-level executors, translating the skill chain from the task planner into concrete actions and sequentially interacting with the environment. 
Policy models in the proposed baseline share a common architecture comprising modality encoders~\citep{chen_soundspaces_eccv20,max_dinov2_tmlr24}, a temporal module, and actor-critic heads. 
The temporal module and actor-critic heads follow prior work~\citep{chen_soundspaces_eccv20, gu_habmm_iclr23}, and the modality encoders are tailored to the specific observation space of different skills.
% [ZZ-40] 原句 "maintain a consistent structure following prior works" 中式英语；且 "heads/modules" 前后不一致
For the Navigate skill, the observation space includes sound, head RGB, and head depth.
% \haonote{In the single-event setting, sound observations are converted into binaural spectrograms and processed by a vanilla CNN encoder~\citep{chen_soundspaces_eccv20}. 
% In the multi-event setting, sound observations are also transformed into binaural spectrograms. An additional location predictor~\cite{chen_semantic_cvpr21} is introduced for sound source location prediction.}
In single-event and multi-event settings, sound observations are converted into binaural spectrograms.
The navigation policy uses a vanilla CNN encoder~\citep{chen_soundspaces_eccv20}, while multi-event navigation additionally incorporates a location predictor~\citep{chen_semantic_cvpr21} for sound source localization.
Manipulation skills include {Pick}, {Place}, {Open Door}, {Open Refrigerator}, Open Washing Machine and so on. The observation spaces consist of head and arm depth and are processed by a CNN encoder~\citep{chen_soundspaces_eccv20,gu_habmm_iclr23}.

\section{Experiment}
%\subsection{Implementation}
\noindent\textbf{Implementation Detail.} 
% We optimize all policy models using the Proximal Policy Optimization (PPO) algorithm~\citep{schulman_ppo_arxiv17} on two NVIDIA 4090 GPUs. 
% The training process spans $5 \times 10^5$ environment steps. 
% Structurally, the Actor and Critic networks are Multi-Layer Perceptrons (MLPs) with a hidden dimension of 512. 
% We utilize the Adam optimizer~\citep{kingma_adam_arxiv14} with an initial learning rate of $3 \times 10^{-4}$, a linear decay schedule, a weight decay of $1 \times 10^{-5}$, and a maximum gradient norm of 0.5. 
% Regarding PPO hyperparameters, we set the discount factor to 0.99, the GAE parameter to 0.95, and the clipping range to 0.2. 
% The update phase involves 2 epochs with 2 mini-batches, using a rollout length of 128 steps. 
% The value loss coefficient is set to 0.5, and the entropy coefficient is 0.01. 
% 
We optimize Actor-Critic policies using PPO~\citep{schulman_ppo_arxiv17} on two NVIDIA 4090 GPUs for $5 \times 10^5$ environment steps. Optimization relies on Adam~\citep{kingma_adam_arxiv14} with an initial learning rate of $3 \times 10^{-4}$ (linear decay), a weight decay of $1 \times 10^{-5}$, and a 0.5 maximum gradient norm. Key PPO hyperparameters include a 128-step rollout, 2 update epochs with 2 mini-batches, a 0.99 discount factor, a 0.95 GAE parameter, and a 0.2 clipping range. The value and entropy loss coefficients are set to 0.5 and 0.01, respectively.
The skill-specific training rewards, detailed execution and termination criteria are provided in the supplementary material.

\noindent\textbf{Baseline.} 
% We train low-level policy models individually and compose them with varying task planners, establishing two categories of baselines.
% First, the {Oracle} baseline employs a privileged task planner that provides the ground-truth skill chains, serving as the performance upper bound of task planning.
% Second, for {practical implementations}, we evaluate two types of task planners:
% (1) {Zero-shot Omni-MLLMs}: To evaluate audio-visual reasoning capabilities, we integrate mainstream, training-free models, including {Qwen2.5-Omni} (3B, 7B)~\citep{xu_qwen2_arxiv25}, MiniCPM-o-4.5~\citep{cui_minicpm_arxiv26} and {Qwen3-Omni-30B-Instruct}~\citep{yang_qwen3_arxiv25}. We refer to these baselines directly by their model names.
% (2) {Supervised Setting-Specific Planner (SSP)}: We introduce a lightweight network explicitly trained on initial observations to predict target sound categories and map them to designated skill chains. To accommodate different settings, SSP processes single-channel log-Mel features and head RGB for single-event tasks, and priority-conditioned two-channel features (without RGB) for multi-event tasks.
% 
We combine individually trained low-level policies with various task planners to compose three baselines.
(1) \textbf{Oracle}: Uses ground-truth skill chains, serving as the planning upper bound.
(2) \textbf{Zero-shot Omni-MLLMs}: We evaluate training-free models including Qwen2.5-Omni (3B, 7B)~\citep{xu_qwen2_arxiv25}, MiniCPM-o-4.5~\citep{cui_minicpm_arxiv26}, and Qwen3-Omni-30B-Instruct~\citep{yang_qwen3_arxiv25}.
(3) \textbf{Supervised Setting-Specific Planner (SSP)}: A lightweight network is trained to map initial observations to skill chains via sound category prediction.
\begin{figure*}[!t]
  \centering
  \vspace{-.1in}
  % ================= 第一行 =================
  % (a) 表格
  \begin{subfigure}[b]{0.49\textwidth}
    \centering
    \renewcommand{\arraystretch}{0.98}
    % 使用 \raisebox{-3.5ex} 将整个表格向下推移，抵消右图 X 轴文字的高度，使其底线直接与右图坐标轴对齐。
    % 若偏差，请微调这里的 -3.5ex (如 -3ex 或 -4ex)
    \raisebox{9ex}{%
        \resizebox{\linewidth}{!}{
            \begin{tabular}{ccccccc}
                \shline
                \multirow{3}{*}[-1ex]{Method} & \multicolumn{4}{c}{Single-event setting} & \multicolumn{2}{c}{\multirow{2}{*}[-.5ex]{Multi-event setting}} \\
                \cmidrule(lr){2-5}
                & \multicolumn{2}{c}{Object Relocation} & \multicolumn{2}{c}{State Transition} & \multicolumn{2}{c}{} \\
                \cmidrule(lr){2-3} \cmidrule(lr){4-5} \cmidrule(lr){6-7}
                & Planning ($\uparrow$) & Completion ($\uparrow$) & Planning ($\uparrow$) & Completion ($\uparrow$) & Planning ($\uparrow$) & Completion ($\uparrow$) \\
                \hline
                Oracle & 100.00 & 71.78 & 100.00 & 28.84 & 100.00 & 11.25 \\
                Oracle$^*$ & 100.00 & 86.13 & 100.00 & 30.38 & 100.00 & 19.00 \\
                \hline
                \hline
                Qwen2.5-Omni-3B & 1.46 & 0.97 & 12.29 & 4.10 & 26.25 & 2.50 \\
                Qwen2.5-Omni-7B & 2.68 & 2.19 & 31.91 & 9.39 & 44.50 & 3.25 \\
                MiniCPM-o-4.5 & 13.38 & 9.25 & \textitbf{41.47} & 12.12 & \textitbf{50.00} & \textitbf{5.00} \\
                Qwen3-Omni-30B & \textitbf{88.56} & \textitbf{63.50} & 39.42 & \textitbf{12.29} & 19.25 & 3.00 \\
                SSP & \textbf{99.03} & \textbf{71.05} & \textbf{99.49} & \textbf{28.67} & \textbf{69.50} & \textbf{7.75} \\
                \hline
                \hline
                Qwen2.5-Omni-3B$^*$ & 1.46 & 1.22 & 12.29 & 3.92 & 26.25 & 2.50 \\
                Qwen2.5-Omni-7B$^*$ & 2.68 & 2.43 & 31.91 & 10.07 & 44.50 & 5.25 \\
                MiniCPM-o-4.5$^*$ & 13.38 & 11.19 & \textitbf{41.47} & \textitbf{12.97} & \textitbf{50.00} & \textitbf{8.75} \\
                Qwen3-Omni-30B$^*$ & \textitbf{88.56} & \textitbf{75.43} & 39.42 & 11.95 & 19.25 & 4.25 \\
                SSP$^*$ & \textbf{99.03} & \textbf{85.16} & \textbf{99.49} & \textbf{30.20} & \textbf{69.50} & \textbf{13.25} \\
                \shline
            \end{tabular}
        }
    }
    
    \subcaption{}
    \label{fig:sota_overall}
  \end{subfigure}\hfill
  % (b) 第一张图
  \begin{subfigure}[b]{0.49\textwidth}
    \centering
    \raisebox{-1.5ex}{\includegraphics[width=0.95\linewidth]{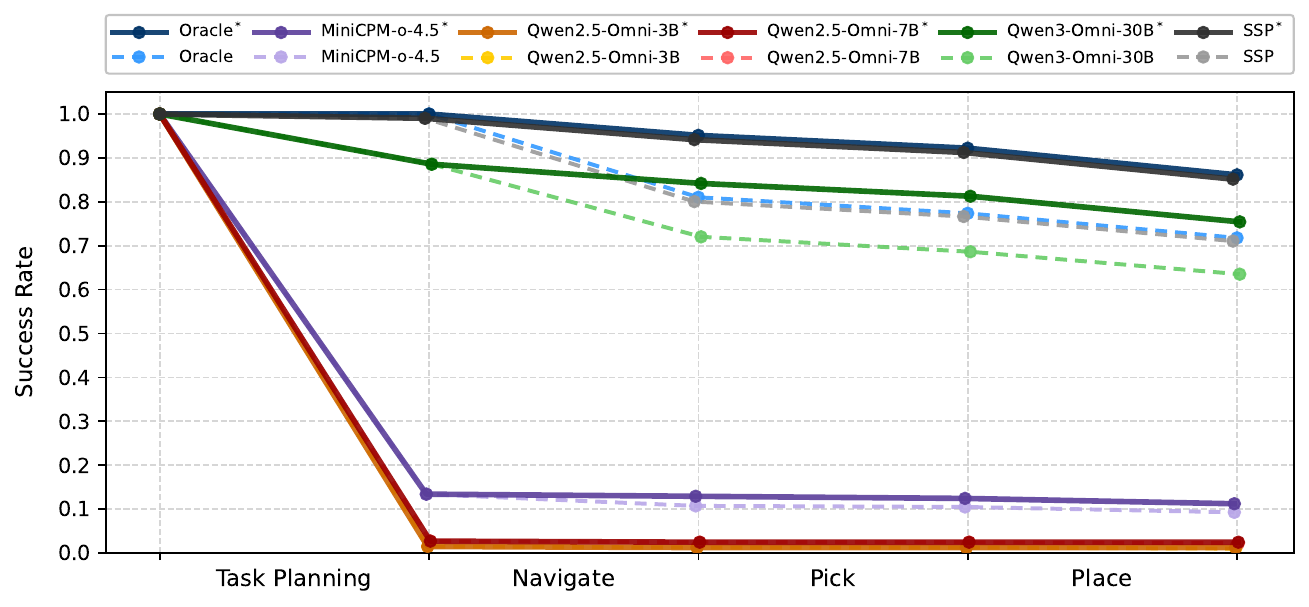}}
    
    \subcaption{}
    \label{fig:pick_place}
  \end{subfigure}\hfill

  % \vspace{1ex} % 行间距

  % ================= 第二行 =================
  % (c) 第三张图 
  \begin{subfigure}[b]{0.48\textwidth}
    \centering
    % 1. 将 width 稍微调小 (例如 0.92\linewidth)，使左图等比例缩小，从而降低网格的绝对高度。
    % 2. 使用 \raisebox 向上平移(例如 0.4cm)，填补底部因为文字较少而产生的视觉高度差。
    \raisebox{0.08cm}{\includegraphics[width=0.92\linewidth]{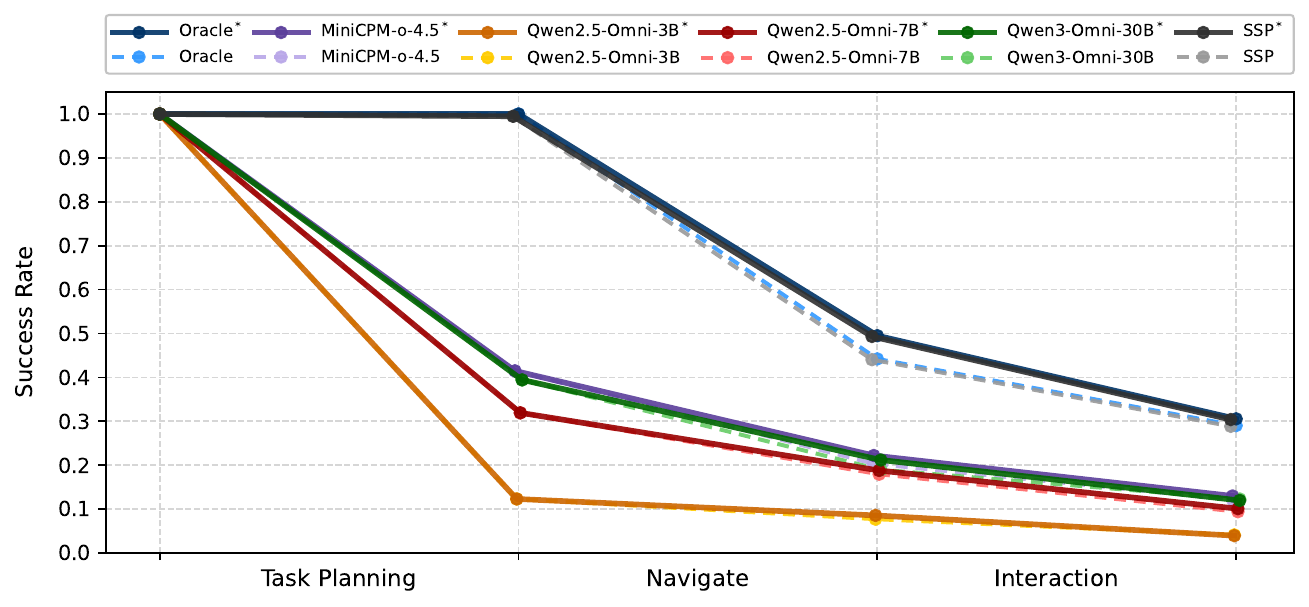}}
    \subcaption{}
    \label{fig:open_door}
  \end{subfigure}\hfill
  % (d) 第四张图
  \begin{subfigure}[b]{0.48\textwidth}
    \centering
    % 右图作为基准，保持 \linewidth 不变。
    \includegraphics[width=\linewidth]{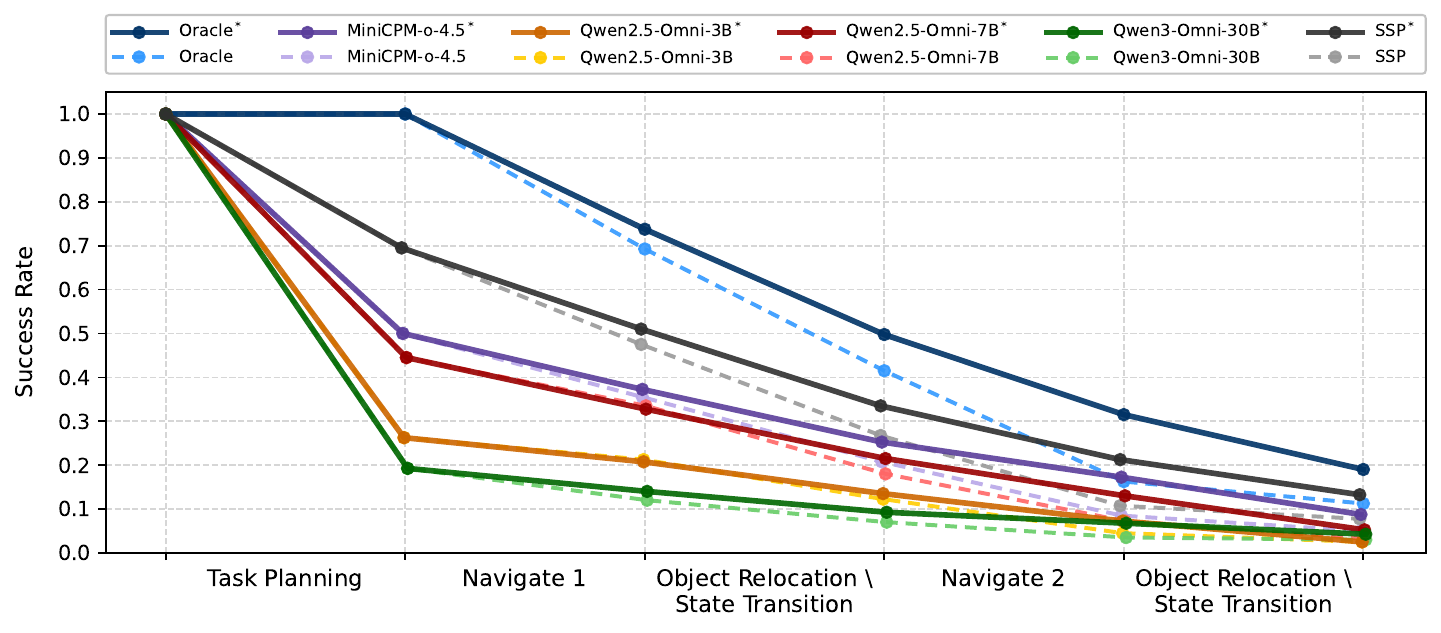}
    \subcaption{}
    \label{fig:2sound}
  \end{subfigure}
  
  \vspace{-0.1in}
\caption{
{Results on STMM-12K.} 
{(a) Success rates of task planning and skill completion.}
Fully-supervised SSP achieves the best performance in both settings. %Meanwhile, MiniCPM-o-4.5 and Qwen3-Omni-30B yield competitive results in a zero-shot manner.
\textbf{Best} and \textitbf{second best}.
{Progressive completion success rate of (b)~object relocation (single-event), (c)~state transition (single-event), and (d)~multi-event setting.} 
% {(c) Progressive completion success rate of state transition (single-event).} 
% {(d) Progressive completion success rate in the multi-event setting.} 
$^*$ denotes the target position that is provided for the navigation policy model. 
}
\label{fig:sota}
\vspace{-0.1in}
\end{figure*}

\begin{figure*}[!t]
    \centering
    \includegraphics[width=0.75\linewidth]{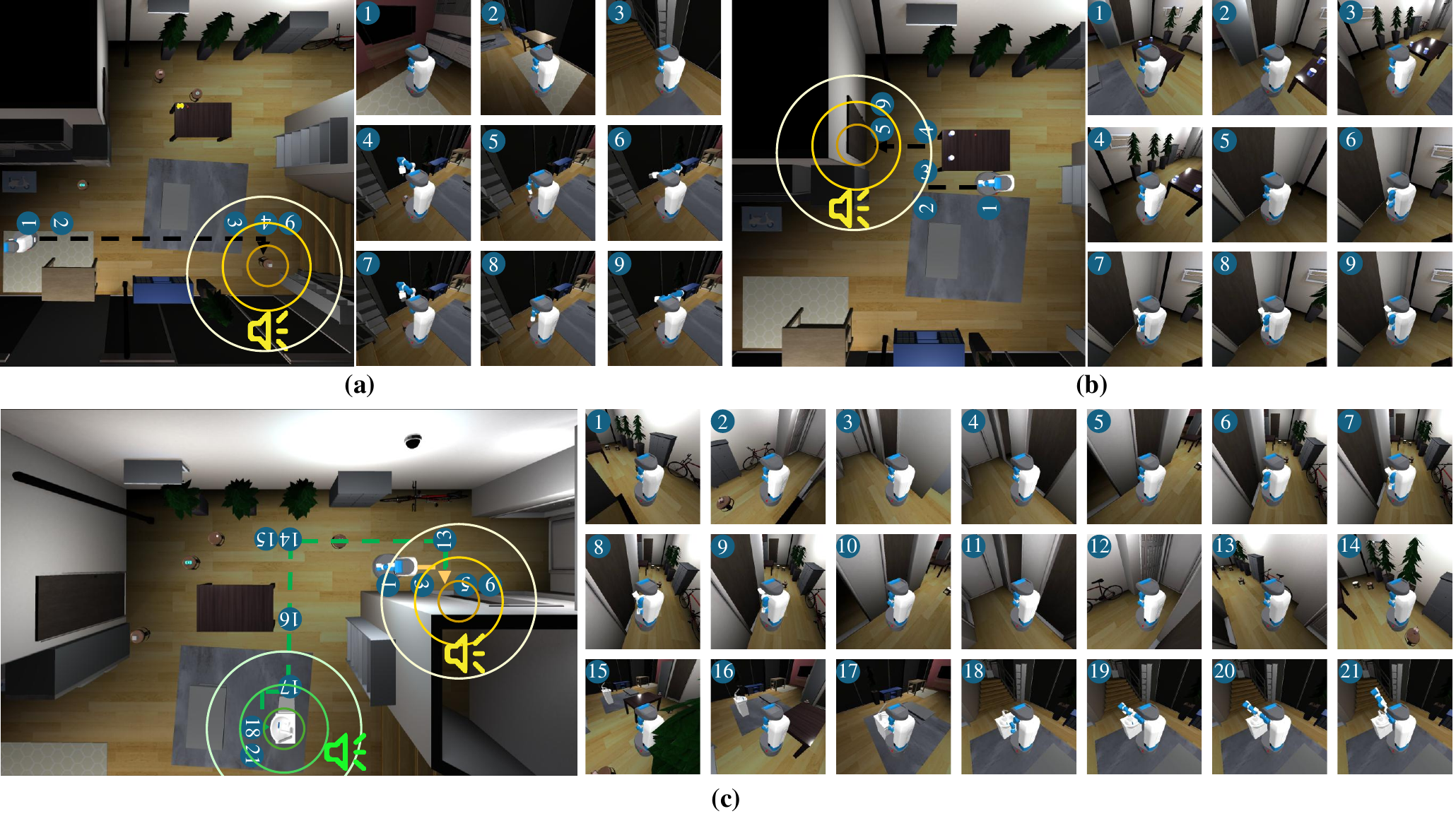}
    \vspace{-.1in}
    \caption{
    {Qualitative Visualizations of Task Execution.} 
Execution trajectories for (a) {Object Relocation}, (b) {State Transition}, and (c) {Multi-event Setting}. For each task, the {left panel} maps robot paths and sound sources from a top-down view, while the {right panel} shows third-person keyframes. Multi-event Setting requires sequential navigation and interaction with two distinct sound sources.
}
    \label{fig:qa}
    \vspace{-.2in}
\end{figure*}

\subsection{Result}
\noindent\textbf{Quantitative Results of Object Relocation in Single-Event setting.}
In~\reffig{fig:sota_overall} and~\reffig{fig:pick_place}, the {Oracle$^*$} baseline, which utilizes ground-truth planning and target position, achieves a completion success rate of 86.13\%.
Regarding task planning, Qwen3-Omni-30B achieves the highest accuracy of 88.56\%, surpassing other zero-shot planners.
Supervised Setting-Specific Planner (SSP) achieves a 99.03\% planning rate and a 71.05\% completion success rate. 

\noindent\textbf{Quantitative Results for State Transition in the Single-Event Setting.} 
\reffig{fig:sota_overall} and \reffig{fig:open_door} show the task planning and completion success rates.
The Oracle$^*$ baseline (where $^*$ denotes adopting the target position) receives a completion success rate of 30.38\%. SSP closely approaches this with 99.49\% planning and 28.67\% completion success rates.
Among zero-shot Omni-MLLMs, MiniCPM-o-4.5 leads in planning (41.47\%), outperforming Qwen2.5-Omni models and followed closely by Qwen3-Omni-30B (39.42\%). This translates to completion rates of 12.12\% and 12.29\% for MiniCPM-o-4.5 and Qwen3-Omni-30B, respectively. Current zero-shot planners lag behind specialized supervised models in reliable task planning.

\noindent\textbf{Quantitative Results of Multi-event Setting.}
We report the multi-event results in \reffig{fig:sota_overall}.
Among zero-shot task planners, MiniCPM-o-4.5 achieves the highest task planning accuracy, followed by Qwen2.5-Omni-7B.
The supervised Setting-Specific Planner (SSP) outperforms all zero-shot planners.
Oracle$^*$ and SSP achieve completion success rates of 19.00\% and 7.75\%, respectively.
We show the progressive completion success rate in~\reffig{fig:2sound}.
Under interference from the second sound source, Oracle$^*$ and SSP complete the first event with success rates of 49.75\% and 26.75\%, respectively.
% orcale^$*$ first event的成功率是 49.75\%
% SSP first event的成功率是26.75\% 
These results highlight the difficulty of the multi-event setting, where concurrent acoustic interference and cascading errors across sequential skills hinder completion (see~\refsection{sec:task}).

\noindent\textbf{Qualitative Result.}
We visualize executions of the robot in single-event and multi-event settings (see~\reffig{fig:qa}).
In the object relocation (single-event), the robot successfully navigates to the target and executes the {Pick} and {Place} skills (see~\reffig{fig:qa}a).
In the state transition (single-event), the robot accurately navigates to the sound-emitting door and completes the {Open Door} skill (see~\reffig{fig:qa}b).
The multi-event setting involves two simultaneous sound sources: \textcolor{stmmdoorbell}{a primary source} and \textcolor{stmmalarm}{a secondary source} (see~\reffig{fig:qa}c).
Following the specified event priority, the robot first interacts with the \textcolor{stmmdoorbell}{primary source} despite interference from the \textcolor{stmmalarm}{secondary} one and then interacts with the secondary source. 
More visualizations can be seen in the supplementary material.

\begin{table*}[!t]
% \vspace{-.1in}
  \centering
  \footnotesize 
  \setlength{\tabcolsep}{3pt}
  \caption{
  {Ablation Study.}
  % We investigate the impact of input modalities across task planning and representative policy models. S.R. denotes success rate.
  {(a)} {Effect of sound modality on navigate policy model.} Cat. and Tgt. Pos. denote the category embedding and target position of the sound source, respectively.
  {(b)} {Task Planning.} Combining Head RGB and sound is the best.
  {(c)} {Policy model for Navigate (Visual Modality).} Using Head RGB and Head Depth yields the highest.
  {(d)} {Policy model for Pick and Open Washing Machine~(O.W.M.).} Combining Head and Arm Depth outperforms single-view.
  }
  \label{tab:ablation_all}
  \vspace{-0.15in}

  % 第一个表：Audio Modality (nav_audio)
  \begin{subtable}[t]{0.27\linewidth} 
  \centering
  \caption{}
    \resizebox{\linewidth}{!}{%
    \begin{tabular}{ccccc}
      \shline
      \multirow{2}[4]{*}{Model} & \multicolumn{3}{c}{Input Modality} & \multirow{2}[2]{*}{S.R. ($\uparrow$)} \\
      \cmidrule(lr){2-4}
            & Sound & Cat. & Tgt. Pos. &  \\
      \hline
      \Rnum{1}      & \textcolor{red}{\ding{55}}      & \textcolor{red}{\ding{55}}      & \textcolor{red}{\ding{55}} & 0.00 \\
      \Rnum{2}      & \textcolor{red}{\ding{55}}      & \textcolor{nvidiagreen}{\ding{51}}      & \textcolor{red}{\ding{55}} & 0.00 \\
      \Rnum{3}      & \textcolor{nvidiagreen}{\ding{51}}      & \textcolor{red}{\ding{55}}      & \textcolor{red}{\ding{55}} & 81.02 \\
      \Rnum{4}      & \textcolor{red}{\ding{55}}      & \textcolor{red}{\ding{55}}      & \textcolor{nvidiagreen}{\ding{51}} & 93.43 \\
      \Rnum{5}      & \textcolor{nvidiagreen}{\ding{51}}      & \textcolor{red}{\ding{55}}      & \textcolor{nvidiagreen}{\ding{51}} & \textbf{95.13} \\
      \shline
    \end{tabular}%
    }
  \label{tab:ablation_nav_audio}
  \end{subtable}%
  \hfill 
  % 第二个表：Task Planning (nav_modality)
  \begin{subtable}[t]{0.19\linewidth} 
  \centering
  \caption{}
    \resizebox{\linewidth}{!}{%
    \begin{tabular}{cccc}
      \shline
      \multirow{2}[4]{*}[1ex]{Model} & \multicolumn{2}{c}{Input Modality} & \multirow{2}[4]{*}[1ex]{S.R. ($\uparrow$)} \\
\cmidrule{2-3}          & Head RGB & Sound &  \\
      \hline
      \Rnum{1}     & \textcolor{nvidiagreen}{\ding{51}}      & \textcolor{red}{\ding{55}}      & 52.31 \\
      \Rnum{2}     & \textcolor{red}{\ding{55}}      & \textcolor{nvidiagreen}{\ding{51}}      & 98.30 \\
      \Rnum{3}     & \textcolor{nvidiagreen}{\ding{51}}      & \textcolor{nvidiagreen}{\ding{51}}      & \textbf{99.03} \\
      \shline
    \end{tabular}%
    }
  \label{tab:ablation_nav_modality}
  \end{subtable}%
  \hfill 
  % 第三个表：Navigate (nav)
  \begin{subtable}[t]{0.22\linewidth} 
    \centering
    \caption{}
    \resizebox{\linewidth}{!}{%
      \begin{tabular}{cccc}
      \shline
      \multirow{2}[2]{*}{Model} & \multicolumn{2}{c}{Input Modality} & \multirow{2}[2]{*}{S.R. ($\uparrow$)} \\
      \cmidrule(lr){2-3}
            & Head RGB & Head Depth &  \\
      \hline
      \Rnum{1}      & \textcolor{red}{\ding{55}}      & \textcolor{nvidiagreen}{\ding{51}}      & 72.74  \\
      \Rnum{2}      & \textcolor{nvidiagreen}{\ding{51}}      & \textcolor{red}{\ding{55}}      & 78.58  \\
      \Rnum{3}      & \textcolor{nvidiagreen}{\ding{51}}      & \textcolor{nvidiagreen}{\ding{51}}      & \textbf{81.02} \\
      \shline
      \end{tabular}%
    }
    \label{tab:ablation_nav}
  \end{subtable}%
  \hfill 
  % 第四个表：Pick and open_washing_machine
  \begin{subtable}[t]{0.28\linewidth} 
    \centering
    \caption{}
    \resizebox{\linewidth}{!}{%
      \begin{tabular}{ccccc}
    \shline
    \multirow{2}[4]{*}[1ex]{Model} & \multicolumn{2}{c}{Input Modality} & \multicolumn{2}{c}{S.R. ($\uparrow$)} \\
\cmidrule(lr){2-3} \cmidrule(lr){4-5}          & Head Depth & Arm Depth & Pick  & O.W.M. \\
    \hline
    \Rnum{1}     & \textcolor{nvidiagreen}{\ding{51}}      & \textcolor{red}{\ding{55}}      & 92.70  & 68.91  \\
    \Rnum{2}     & \textcolor{red}{\ding{55}}      & \textcolor{nvidiagreen}{\ding{51}}      & 92.94  & 56.75  \\
    \Rnum{3}     & \textcolor{nvidiagreen}{\ding{51}}      & \textcolor{nvidiagreen}{\ding{51}}      & \textbf{96.59}  & \textbf{70.27}  \\
    \shline
    \end{tabular}%
    }
    \label{tab:ablation_pick_open_washing_machine}
  \end{subtable}
\vspace{-.25in}
\end{table*}

\subsection{Ablation Study and Further Discussion}

\noindent\textbf{Effect of Sound Modality for Navigate Skill.} 
% [ZZ-48] 这一段被完整写了两遍（下面用 % 注释掉的旧版 + 未注释的精简版）。
%         已删除旧版，仅保留一份，并把两份里更硬的数字解读合并。
% 说明：删除线内 \Rnum 换成字面数字（ulem 不接受 \uppercase），交叉引用用字面 "Table 1" 占位。
% \zzdel{Table~1 presents the impact of different input modalities for the navigate skill. Sound signals are converted into binaural spectrograms to provide implicit position information. Models lacking any spatial information (Model 1 and Model 2) completely fail to navigate, yielding a 0.00\% success rate. Relying solely on sound signals achieves an 81.02\% success rate (Model 3). Using the explicit target position directly reaches 93.43\% (Model 4). Combining both sound and explicit target positions yields the best performance at 95.13\% (Model 5). Overall, these results indicate two key findings. First, spatial information of the sound source is crucial. Second, as an environmental signal, sound provides effective position information and is complementary to the explicit target position.}
% 
We ablate the input modalities of the navigation skill(see~\reftab{tab:ablation_nav_audio}), where sound is processed into binaural spectrograms for implicit positioning. Models lacking spatial data fail completely (Model~\Rnum{1} and \Rnum{2}, 0.00\%). Success scales up when using sound only (Model~\Rnum{3}, 81.02\%) or explicit target positions (Model~\Rnum{4}, 93.43\%). Combining both achieves the optimal 95.13\% (Model~\Rnum{5}). This confirms that spatial information is indispensable and sound serves as an effective positional cue complementary to explicit coordinates.

\noindent\textbf{Effect of input Modality for Task Planning.}
As shown in~\reftab{tab:ablation_nav_modality}, we ablate the input modalities used for task planning in object relocation. We utilize Setting-Specific Planner (SSP) as the baseline.
Relying exclusively on Head RGB yields the lowest success rate at 52.31\%. Utilizing only sound signals substantially elevates the performance to 98.30\%. Integrating both Head RGB and sound attains the optimal result of 99.03\%. These results show the importance of auditory information in task planning, while visual input from Head RGB is supplementary.

\noindent\textbf{Effect of Visual Modality for Navigate Skill.} 
We evaluate the impact of different visual modalities, including head RGB and head depth for the Navigate skill (see~\reftab{tab:ablation_nav}).
Head-depth-only input performs worst, because depth carries no semantic texture and cannot be aligned with sound semantics; Head RGB alone performs better; and combining Head RGB with Head depth gives the best overall performance.
% This combination is effective because the modalities are complementary. RGB provides semantic context to align with sound sources, while depth offers precise geometric distance information.
This combination leverages complementary modalities: RGB provides sound-aligned semantics, while depth yields precise geometry.

\noindent\textbf{Effect of Input Modality for Pick Skill and Open Washing Machine Skill.} 
% We investigate the impact of depth modalities on the {Pick} skill and Open Washing Machine Skill in~\reftab{tab:ablation_pick_open_washing_machine}, comparing the efficacy of head depth and arm depth.
% For the Pick skill, using head depth alone yields the lowest success rate of 92.70\%, primarily because the target object often falls outside the head camera's optimal viewing range during manipulation.
% In contrast, the model relying solely on arm depth achieves a higher success rate of 92.94\%, as the arm depth can directly observe the target when aligned.
% We observe that integrating both depth sources results in the best performance (96.35\%).
% A similar trend appears in the Open Washing Machine Skill, where combining both depth views reaches the optimal result of 70.27\%, significantly outperforming either single-modality setup.
% The two views are complementary: the arm depth provides precise object-to-gripper distance and local affordances, while the head depth offers a global context of the gripper and object.
In~\reftab{tab:ablation_pick_open_washing_machine}, integrating head and arm depth maximizes performance for both Pick (96.59\%) and Open Washing Machine (70.27\%) skills. Single-view input underperforms: for Pick, head depth alone (92.70\%) suffers from manipulation blind spots, while arm depth (92.94\%) lacks broader spatial awareness. The two views are complementary, fusing precise local geometry from the arm with global context from the head.

\section{Conclusion}
In this work, we introduce Sound-Triggered Mobile Manipulation (STMM), a new paradigm in which robots respond to sound events by physically interacting with the corresponding sound-emitting objects. 
To support this paradigm, we develop {Habitat-Echo}, a simulation platform integrating sound rendering with physical interaction, and establish a hierarchical baseline comprising a high-level task planner and low-level policies. 
Experiments show that the baseline localizes and interacts with sound sources in the single-event setting.
In the multi-event setting, the baseline shows the feasibility of sequential interaction with multiple sound sources according to a prescribed priority, providing a starting point for further research on STMM.

\section{AI use statement}
In this work, we used generative AI tools (\eg, ChatGPT) to edit the research paper to improve readability, specifically for grammar and spelling checks. We have not used generative AI tools for creating scientific figures or any plagiarism. We have carefully reviewed all AI-assisted text revisions to ensure their accuracy. We take full responsibility for the final content of this work.

\bibliography{iclr2027_conference}
\bibliographystyle{iclr2027_conference}

\clearpage
\appendix
\section{Appendix}
\newcommand{\apptocline}[3][0em]{%
  \par\noindent\hspace*{#1}%
  \hyperref[#2]{\makebox[2.2em][l]{\ref*{#2}}#3}%          % 字母 + 标题（可点击）
  \nobreak\dotfill\ \makebox[1.8em][r]{\pageref{#2}}\par   % 点线 + 右对齐页码
  \vspace{2pt}%
}

\begin{center}{\large\scshape Contents}\end{center}
\vspace{-0.4em}\hrule\vspace{0.6em}
{\small
\apptocline{sec:additional_related_work}{Additional Related Work}
\apptocline{sec:prompt}{Prompt for Task Planning}
\apptocline{sec:ablation}{Additional Ablation Study}
\apptocline{sec:add_qa}{Additional Qualitative Results}
\apptocline{sec:skill_space}{Execution Space and Termination Criteria of Skills}
\apptocline{sec:performance}{Detailed Numerical Results on STMM-12k}
\apptocline{sec:policy_detail}{Details of the Low-Level Policy and the Setting-Specific Planner (SSP)}
\apptocline{sec:rir}{RIRs for Sound Rendering}
\apptocline{sec:limitation}{Limitation}

}
\vspace{0.4em}\hrule

\subsection{Additional Related Work}
\label{sec:additional_related_work}
\noindent\textbf{Mobile Manipulation.} 
Mobile manipulation extends the robot's workspace from confined tabletops to large-scale environments, requiring the synergy of mobility and interaction. 
Existing methods fall into two primary streams.
The first stream focuses on {object rearrangement} in structured environments. 
\citep{andrew_habitat_nips21} and \citep{shukla_maniskill_iclr24} develop high-fidelity simulators to support complex physical training and evaluation. 
Building on such platforms, \citep{gu_habmm_iclr23} proposes a hierarchical framework sequencing navigation and interaction skills, while \citep{kant_housekeep_eccv22} integrates commonsense reasoning to infer logical object placements.
The second stream targets {open-vocabulary manipulation} via VLMs. 
\citep{zitkovich_rt2_CRL23} transfers web-scale knowledge to robotic control for generalization. 
To handle unseen objects, \citep{yenamandra_homerobot_arxiv23} establishes a rigorous benchmark, and \citep{liu_ok_arxiv24} achieves zero-shot performance by combining VLMs with navigation primitives.
However, these methods remain predominantly instruction-driven. Relying on predefined textual commands limits the ability of robots to automatically process and actively respond to dynamic events in the environment.

\noindent\textbf{Task Planning.} 
Recent LLM advancements greatly broaden the scope of task planning. 
Specifically, Tool-Planner~\citep{liu_tool_iclr25} minimizes error propagation in tool usage by effectively grouping and selecting APIs. 
In algorithmic planning, \citep{katz_tos_nips24} boosts planning efficiency while maintaining logical rigor through optimized tree-based methods. 
For web agents, \citep{wang_ecomscriptbench_acl25} proposes a benchmark evaluating LLMs' ability to translate user intentions into step-by-step product associations.
Task planning in robotic manipulation bridges high-level reasoning with low-level execution.
Early approaches~\citep{kaelbling_hierarchical_icra11,toussaint_logic_ijcai15,srivastava_combined_icra14} primarily relied on predefined symbolic representations for continuous control. Recent research leverages the extensive world knowledge within LLMs and VLMs for planning and can be broadly categorized into two streams
The first stream focuses on task decomposition.
Wake~\etal~\citep{wake_gpt_ral24} utilize GPT-4V~\citep{openai_gpt4v_23} to ground spatial-temporal information within the demonstration video and analyze affordances for plan generation.
% Similarly, Kachaev~\etal~\citep{kachaev_mind_arxiv25} employ VLMs to decompose instructions into subtask sequences. 
RoboBrain~\citep{ji_robobrain_cvpr25} unifies planning with affordance perception and trajectory prediction within a single model, enhancing manipulation capabilities.
The second stream emphasizes feedback-based refinement.
Zhi~\etal~\citep{zhi_closed_icra25} optimize code-based plans using execution feedback, and Feng~\etal~\citep{feng_reflective_arxiv25} improve plans by anticipating future states.
However, these works primarily rely on visual and textual inputs, neglecting the importance of sound signals.

\noindent\textbf{Omni-Modal Large Language Models (Omni-MLLMs)}
Recent advancements in Omni-MLLMs have transitioned from cascaded pipelines to end-to-end architectures, broadly categorized into three paradigms:
\textit{Adapted End-to-End Speech LLMs:} Models like Qwen2.5-Omni \citep{xu_qwen2_arxiv25} and Llama-Omni \citep{fang2025llama} align robust text backbones with speech encoders. By bypassing intermediate text translation, they retain critical acoustic nuances (e.g., tone, emotion) and significantly reduce response latency.
\textit{Streaming Omni Models:} Architectures such as MiniCPM-o \citep{cui_minicpm_arxiv26} and Moshi \citep{defossez2024moshi} are designed for full-duplex processing of parallel streams. They excel in reactive environments by dynamically handling continuous auditory inputs and sudden user interruptions.
\textit{Native Omni-Modal Foundation Models:} Qwen3-Omni \citep{yang_qwen3_arxiv25} tokenizes audio, text, and visual data into a unified representation space from the very first layer, maximizing complex cross-modal reasoning without internal translation bottlenecks.
Current omni-MLLMs excel at processing human speech. However, omni-MLLMs show limited generalization to natural environmental sounds, presenting a significant challenge for comprehensive task planning in STMM.

\subsection{Prompt for Task Planning}
\label{sec:prompt}
To ensure a fair comparison, we utilize a standardized prompt structure across all task planners. 
The input consists of two components: a {system prompt} and a {user prompt}. 
The system prompt defines the detailed rules and logic for task planning, while the user prompt provides the environmental context, including visual and acoustic observations.

\subsubsection{Prompt for Task Planning in the Single-event Setting.} 
The system prompt for the single-event setting is structured into four components. 
The first component defines the strictly allowed skill vocabulary. 
The second establishes the logical mapping between sound and the corresponding skill chains. 
The third specifies the planning constraints and the mandatory JSON output format. 
Finally, the fourth component provides a structural template for the response. 
Complementing this, the user prompt supplies the visual and acoustic observations to ground the planning process.

\begin{tcolorbox}[
    breakable,
    title= (\Rnum{1}) Prompt for Task Planning in the Single-event Setting,
    before skip=10pt,      % 强制设定盒子与上方文字的间距（可以改得更小，如 0pt）
    lines before break=1, % 关键！告诉 latex 即使当前页只能放 1 行，也允许在这里开始盒子
    after skip=34pt, % 盒子与下方文字的间距 (默认可能比较大，您可以设为 6pt 甚至 0pt)
]

\textbf{System Prompt:} You are an expert Embodied AI Task Planner. Your goal is to generate a strictly ordered list of skills based on the input auditory (sound) and visual (head\_rgb) observations.

\textbf{1. ALLOWED SKILL VOCABULARY:}
You are strictly limited to ONLY the following skills. Do not invent new words or change spellings: [``nav", ``pick", ``place", ``open\_door", ``close\_sink", ``open\_air\_fryer", ``open\_dishwasher", ``open\_microwave", ``open\_refrigerator", ``open\_washing\_machine"].

\textbf{2. LOGIC MAPPING (AUDIO $\rightarrow$ SKILL):} Analyze the input audio sound and classify the source object to determine the skill sequence:
\begin{enumerate}
    \item Category A: Relocation Objects
    \begin{enumerate}
        \item If the sound source is:  Mechanical Alarm, Phone (ringing tone), Furby (toy), Talkie, Switch, Radio, Music Box, Bluetooth speaker, Earbuds, Watch, or Tablet. REQUIRED PLAN: [``nav", ``pick", ``place"]
    \end{enumerate}
    \item Category B: State Transition
    \begin{enumerate}
        \item If the sound source is: Doorbell. REQUIRED PLAN: [``nav", ``open\_door"]
        \item If the sound source is: Water Running. REQUIRED PLAN: [``nav", ``close\_sink"]
        \item If the sound source is: Air Fryer Beeping. REQUIRED PLAN: [``nav", ``open\_air\_fryer"]
        \item If the sound source is: Dishwasher Running. REQUIRED PLAN: [``nav", ``open\_dishwasher"]
        \item If the sound source is: Microwave Beeping. REQUIRED PLAN: [``nav", ``open\_microwave"]
        \item If the sound source is: Refrigerator Alarm. REQUIRED PLAN: [``nav", ``open\_refrigerator"]
        \item If the sound source is: Washing Machine Running. REQUIRED PLAN: [``nav", ``open\_washing\_machine"]
    \end{enumerate}
\end{enumerate}
\textbf{3. CONSTRAINTS and OUTPUT FORMAT:}
\begin{enumerate}
    \item Never return an empty list.
    \item Output MUST be a valid JSON object containing a single key ``plan".
    \item Do not include markdown formatting (like \texttt{```} json), just the raw JSON string.
\end{enumerate}
\textbf{4. OUTPUT FORMAT: } \{``plan": [``skill\_1", ``skill\_2", ...]\}
\newline

\textbf{User Prompt:} 
\begin{enumerate}
    \item \textsc{Image Input}
    \item \textsc{Audio Input}
    \item ``Input observation provided. Based on the audio class and visual context, output the execution plan JSON."
\end{enumerate}

\end{tcolorbox}

\subsubsection{Prompt for Task Planning in the Multi-event Setting.} 
For the multi-event setting, we organize the system prompt into five segments to handle the dual-source complexity.
The initial segment lists the permissible skill vocabulary.
The second details the correspondence between sound categories and their required skill sequences.
Crucially, the third segment enforces {interaction priority}. We explicitly identify the primary sound source, requiring the planner to deduce the secondary source from the mixed audio and generate a sequential execution plan for both.
The fourth and fifth segments impose strict output constraints and provide a valid JSON template to ensure parsing reliability.
Finally, the user prompt provides the model with multimodal context, including visual frames and sound data.
\begin{tcolorbox}[
    breakable,
    title= (\Rnum{2}) Prompt for Task Planning in the Multi-event Setting,
    before skip=10pt,      % 强制设定盒子与上方文字的间距（可以改得更小，如 0pt）
    lines before break=1, % 关键！告诉 latex 即使当前页只能放 1 行，也允许在这里开始盒子
    after skip=34pt, % 盒子与下方文字的间距 (默认可能比较大，您可以设为 6pt 甚至 0pt)
]

\textbf{System Prompt:} You are an expert Embodied AI Task Planner. Your goal is to generate a strictly ordered list of skills based on the input auditory and visual observations. There are exactly two sound sources detected.

\textbf{1. ALLOWED SKILL VOCABULARY (STRICT):}
You are strictly limited to ONLY the following skills: [``nav", ``pick", ``place", ``open\_door", ``close\_sink", ``open\_air\_fryer", ``open\_dishwasher", ``open\_microwave", ``open\_refrigerator", ``open\_washing\_machine"]. Do not invent new words or change spellings.

\textbf{2. LOGIC MAPPING (AUDIO $\rightarrow$ SKILL):} Analyze the input audio sound to determine the action sequence:
\begin{enumerate}
    \item Category A: Relocation Objects
    \begin{enumerate}
        \item If the sound source is:  Mechanical Alarm, Phone (ringing tone), Furby (toy), Talkie, Switch, Radio, Music Box, Bluetooth speaker, Earbuds, Watch, or Tablet. REQUIRED PLAN: [``nav", ``pick", ``place"]
    \end{enumerate}
    \item Category B: State Transition
    \begin{enumerate}
        \item If the sound source is: Doorbell. REQUIRED PLAN: [``nav", ``open\_door"]
        \item If the sound source is: Water Running. REQUIRED PLAN: [``nav", ``close\_sink"]
        \item If the sound source is: Air Fryer Beeping. REQUIRED PLAN: [``nav", ``open\_air\_fryer"]
        \item If the sound source is: Dishwasher Running. REQUIRED PLAN: [``nav", ``open\_dishwasher"]
        \item If the sound source is: Microwave Beeping. REQUIRED PLAN: [``nav", ``open\_microwave"]
        \item If the sound source is: Refrigerator Alarm. REQUIRED PLAN: [``nav", ``open\_refrigerator"]
        \item If the sound source is: Washing Machine Running. REQUIRED PLAN: [``nav", ``open\_washing\_machine"]
    \end{enumerate}
\end{enumerate}
\textbf{3. OBJECT MAPPING INSTRUCTION (CRITICAL):}
You are provided with an event-priority specification: One of the sound sources has a high priority. You have to interact with ``\texttt{\{obj\_1\}}" first. Your task is to identify the other sound source from the audio and generate the plan for the whole process. You MUST map the objects exactly as follows:
\begin{enumerate}
    \item Key ``first\_sound": Generate the plan for ``\texttt{\{obj\_1\}}".
    \item Key ``second\_sound": Identify the SECOND sound source from the audio (the one that is NOT ``\texttt{\{obj\_1\}}") and generate its plan.
    \item Do NOT output plans for any other potential objects unless they are detected as the second sound.
\end{enumerate}
\textbf{4. CONSTRAINTS and OUTPUT FORMAT:}
\begin{enumerate}
    \item Output MUST be a valid JSON object containing a single root key ``plan".
    \item The keys inside ``plan" MUST be exactly {``}first\_sound" and ``second\_sound".
    \item Do not include markdown formatting (like \texttt{```} json), just the raw JSON string.
\end{enumerate}
\textbf{5. Expected JSON Structure:} \{
        ``plan": \{
            
            ``first\_sound": [``skill\_1", ``skill\_2", ...],  \# $\leftarrow$ Actions for the object assigned to 'first\_sound'

            ``second\_sound": [``skill\_1", ``skill\_2", ...]  \# $\leftarrow$ Actions for the object assigned to 'second\_sound'
        
        \}
    \}
\newline

\textbf{User Prompt:} 
\begin{enumerate}
    \item \texttt{Image Input}
    \item \texttt{Audio Input}
    \item ``Input observation provided. Based on the audio class and visual context, output the execution plan JSON."
\end{enumerate}

\end{tcolorbox}

\subsection{Additional Ablation Study}
\label{sec:ablation}
We evaluate the impact of different input modalities for the Setting-Specific Planner (SSP) in the multi-event setting (see~\reftab{tab:ablation_task_planning}).
In the multi-event setting, combining head RGB and sound brings more distractions for task planning than sound alone, unlike in the single-event setting.

\begin{comment}
We evaluate the impact of different visual modalities, including head depth, arm depth for the Place and Close Sink policy models (see~\reftab{tab:ablation_place_close_sink}).
The model combining both depth inputs performs best in the Place skill. Specifically, the head depth estimates the arm and gripper pose, while the arm depth provides the distance to the target position. Relying on a single depth input causes a slight performance decline.
Similarly, combining both depth inputs outperforms single-view models, while the arm-depth model performs the worst. 
\begin{table}[h]
    \centering
    \caption{Ablation study on input modalities for the Pick and Close Sink policy models.\todohao{discard?}}
    \small
    % \resizebox{\linewidth}{!}
    {
      \begin{tabular}{ccccc}
    \shline
    \multirow{2}[4]{*}[1ex]{Model} & \multicolumn{2}{c}{Input Modality} & \multicolumn{2}{c}{S.R. ($\uparrow$)} \\
\cmidrule(lr){2-3} \cmidrule(lr){4-5}          & Head Depth & Arm Depth & Place  & Close Sink \\
    \hline
    \Rnum{1}     & \textcolor{nvidiagreen}{\ding{51}}      & \textcolor{red}{\ding{55}}      & 68.46  & 75.00  \\
    \Rnum{2}     & \textcolor{red}{\ding{55}}      & \textcolor{nvidiagreen}{\ding{51}}      & 68.91  & 67.85  \\
    \Rnum{3}     & \textcolor{nvidiagreen}{\ding{51}}      & \textcolor{nvidiagreen}{\ding{51}}      & \textbf{70.27}  & \textbf{77.67}  \\
    \shline
    \end{tabular}}
    \label{tab:ablation_place_close_sink}
  \end{table}      
\end{comment}
% 第二个表：Task Planning (nav_modality)
  \begin{table}[h]
  \centering
  \caption{Ablation study on input modalities for Setting-Specific Planner (SSP) in the multi-event setting.}
    \resizebox{0.4\linewidth}{!}{%
    \begin{tabular}{cccc}
      \shline
      \multirow{2}[4]{*}[1ex]{Model} & \multicolumn{2}{c}{Input Modality} & \multirow{2}[4]{*}[1ex]{S.R. ($\uparrow$)} \\
\cmidrule{2-3}          & Head RGB & Sound &  \\
      \hline
      \Rnum{1}     & \textcolor{nvidiagreen}{\ding{51}}      & \textcolor{red}{\ding{55}}      & 6.25 \\
      \Rnum{2}     & \textcolor{nvidiagreen}{\ding{51}}      & \textcolor{nvidiagreen}{\ding{51}}      & 68.25 \\
      \Rnum{3}     &  \textcolor{red}{\ding{55}}     & \textcolor{nvidiagreen}{\ding{51}}      & \textbf{69.50} \\
      \shline
    \end{tabular}%
    }
  \label{tab:ablation_task_planning}
  \end{table}%

\subsection{Additional Qualitative Result}
\label{sec:add_qa}
\begin{figure*}[!h]
    \centering
    \includegraphics[width=0.99\linewidth]{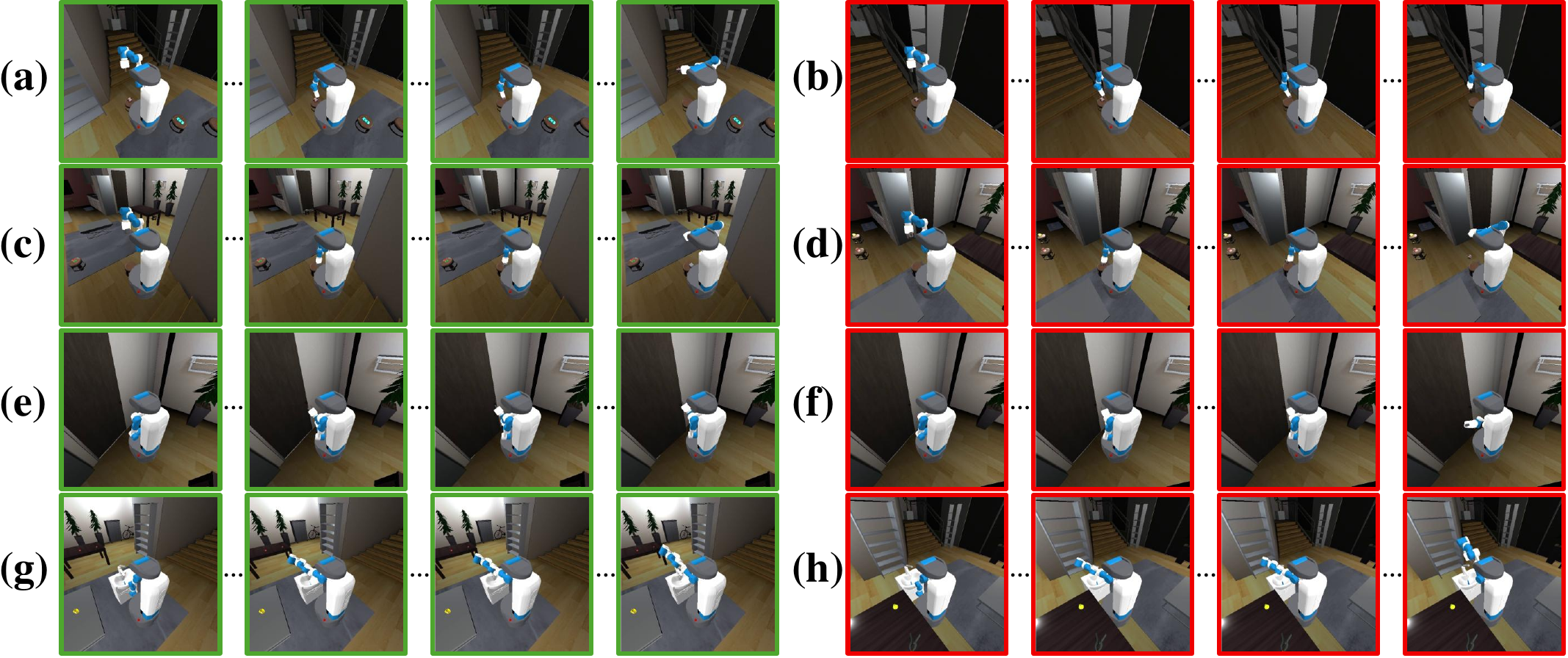}% 中括号中的为调节图片大小
    \caption{
    \emphline{Additional Qualitative Results of Different Skills.} 
    {(a)}, {(b)} \textcolor{nvidiagreen}{Success} and \textcolor{red}{failure} cases of the Pick skill.
    {(c)}, {(d)} \textcolor{nvidiagreen}{Success} and \textcolor{red}{failure} cases of the Place skill.
    {(e)}, {(f)} \textcolor{nvidiagreen}{Success} and \textcolor{red}{failure} cases of the Open Door skill.
    {(g)}, {(h)} \textcolor{nvidiagreen}{Success} and \textcolor{red}{failure} cases of the Close Sink skill.
    }
    \label{fig:qa_supp}
    % \vspace{-.2in}
\end{figure*}
We present qualitative visualization results for the four manipulation skills in~\reffig{fig:qa_supp}. \reffig{fig:qa_supp}a and \reffig{fig:qa_supp}b illustrate the Pick skill. In the success case (a), the robot grasps the object and correctly returns the arm to the resting position, while in the failure case (b), the arm fails to retract to the standard pose despite holding the object. The Place skill is depicted in \reffig{fig:qa_supp}c and d. In \reffig{fig:qa_supp}c, the object is stably placed on the table, whereas in (d), the object is released with an unstable pose and falls off the surface of the small table. \reffig{fig:qa_supp}e and f visualize the Open Door skill. In the success scenario (e), the arm pulls the door open and resets to the resting position, but in the failure scenario \reffig{fig:qa_supp}f, the arm remains extended after opening the door. Finally, \reffig{fig:qa_supp}g and h show the Close Sink skill, where the height of the blue water column visually correlates with the faucet handle angle. In \reffig{fig:qa_supp}g, the faucet is successfully closed, while in \reffig{fig:qa_supp}h, the arm accidentally collides with the handle, causing the opening angle to increase rather than decrease. \textit{More visualizations can be seen in the attached videos.} We provide video demos for different settings (use headphones for a better experience).

\subsection{Execution Space and Termination Criteria of Skill}
\label{sec:skill_space}
Similar to previous work~\cite{gu_habmm_iclr23,shukla_maniskill_iclr24,wang_instruction_aaai25}, the proposed tasks are decomposed into different subtasks. Each subtask is solved by an individual skill. A task is considered completed successfully if and only if all constituent skills are successfully executed in sequence. Skill spaces are shown below.

\noindent\textbf{Navigate.} 
The action space consists of four actions: \texttt{MoveForward}, \texttt{TurnLeft}, \texttt{TurnRight}, and \texttt{Stop}.
Specifically, \texttt{MoveForward} translates the robot by 0.5m along its local forward axis, while rotational actions (\texttt{TurnLeft}, \texttt{TurnRight}) execute $90^\circ$ in-place rotations to the left and right, respectively.
The \texttt{Stop} action terminates the execution of navigation without changing the robot's pose and location.
An episode is considered successful if and only if the robot executes \texttt{Stop} while positioned within a region of {0.5m radius} of the goal.

\noindent\textbf{Pick.}
The robot operates in an {8-DoF continuous action space}. % ($\mathcal{A} \in \mathbb{R}^8$).
The first seven dimensions control the arm's joint angles, driving the end-effector to the target pose via the kinematic chain~\cite{abdelmaksoud_depth_access24}, while the final dimension modulates the gripper state.
A Pick skill is considered successful with two conditions:
(1) {Grasping:} The gripper action is positive while an object is within a {15cm snapping threshold} of the end-effector, triggering a stable grasp.
(2) {Retraction:} The robot effectively holds the object and returns the arm to a {predefined resting pose}, ensuring a smooth transition to subsequent subtasks.

\noindent\textbf{Place.} 
The robot operates in an 8-DoF action space identical to that of the Pick skill.
The success conditions for the Place skill are twofold:
(1) {Releasement:} The gripper opens to release the object, ensuring it settles within 15cm of the target position.
(2) {Retraction:} Upon release, the arm reverts to the {default resting pose}, preparing for the next subtask.

\noindent\textbf{Open Door.} 
Operating within the 8-DoF action space, the robot's end-effector attaches to the handle of the door and drives it along its own functional trajectory.
The skill is considered successful with two conditions:
(1) {Opening:} While connected with the door handle, the robot must drive the door's joint angle to a {specified target threshold}.
(2) {Retraction:} The robot disengages from the door handle and returns the arm to the {resting pose}.

\noindent\textbf{Close Sink.}
Similar to the Open Door skill, the robot utilizes the 8-DoF action space to manipulate the sink's {rotary mechanism}.
The end-effector must grasp the faucet handle and apply torque to rotate it along its functional axis.
Success is determined by two conditions:
(1) {Closure:} While maintaining grasp, the robot must rotate the handle to the zero-degree threshold {within a tolerance of 0.2 rad}.
(2) {Retraction:} Upon completing the rotation, the robot releases the handle and reverts the arm to the {resting pose}.

\noindent\textbf{Open Dishwasher.}
Similar to the Open Door skill, the robot uses the 8-DoF action space to rotate the dishwasher's {bottom-hinged} door about its hinge axis.
The end-effector grasps the outer door panel and drives the door along its downward functional trajectory.
Success is determined by two conditions:
(1) {Opening:} While maintaining grasp, the robot must rotate the door to $0.483\,\mathrm{rad}$ {within a tolerance of $0.05\,\mathrm{rad}$}.
(2) {Retraction:} Upon completing the rotation, the robot releases the door and reverts the arm to the {resting pose}.

\noindent\textbf{Open Microwave.}
Operating within the 8-DoF action space, the robot's end-effector grasps the microwave's {side-hinged} door and rotates it about the hinge.
The skill is considered successful with two conditions:
(1) {Opening:} While connected with the door, the robot must drive the joint angle to $0.507\,\mathrm{rad}$ {within a tolerance of $0.15\,\mathrm{rad}$}.
(2) {Retraction:} The robot disengages from the door and returns the arm to the {resting pose}.

\noindent\textbf{Open Air-Fryer.}
Unlike the hinged-door skills, this skill actuates a {prismatic} drawer in the same 8-DoF action space.
The end-effector grasps the drawer front and translates it along the slide rail.
The skill is considered successful with two conditions:
(1) {Opening:} While maintaining grasp, the robot must pull the drawer to the specified target of $0.073\,\mathrm{m}$ {within a tolerance of $0.025\,\mathrm{m}$}.
(2) {Retraction:} The robot releases the drawer and returns the arm to the {resting pose}.

\noindent\textbf{Open Washing Machine.}
The robot utilizes the 8-DoF action space to open the washing machine's {front-loading circular} door.
The end-effector grasps the outer door panel and applies torque about the hinge.
Success is determined by two conditions:
(1) {Opening:} While maintaining grasp, the robot must rotate the door to $0.507\,\mathrm{rad}$ {within a tolerance of $0.05\,\mathrm{rad}$}.
(2) {Retraction:} Upon completing the rotation, the robot releases the door and reverts the arm to the {resting pose}.

\noindent\textbf{Open Refrigerator.}
Operating within the 8-DoF action space, the robot's end-effector grasps the refrigerator's {right-hand side-hinged} door and drives it along its functional trajectory.
The skill is considered successful with two conditions:
(1) {Opening:} While connected with the door, the robot must drive the joint angle to $0.507\,\mathrm{rad}$ {within a tolerance of $0.05\,\mathrm{rad}$}.
(2) {Retraction:} The robot disengages from the door and returns the arm to the {resting pose}.

\subsection{Detailed Numerical Results on the STMM-12k}
\label{sec:performance}
\subsubsection{Evaluation Metrics}
% progressive success rate是最终的成功率，由以下几项成功率共同决定 1. task planning success rate 2. skill completion success rate
% task planning 决定了要调用的policy model，最终是如何建立progressive success rate
% skill completion 是每个policy model执行skill的成功率，是基于task planning的进一步执行结果，在成功执行skill后，在成功的episode上继续执行下一个skill，直到skill chain 执行完毕，最后的progressive success rate = skill chain 中最后一个skill执行的成功的episode数目 / 总的episode数目
To rigorously evaluate the robustness of policy models in Sound-Triggered Mobile Manipulation (STMM), we introduce the completion success rate (see~\reffig{fig:sota}).
A complete task episode in STMM decomposes into an initial high-level task planning phase followed by a sequence of $K$ low-level skills executed by different policy models, denoted as $S = \{S_1, S_2, \dots, S_K\}$. The evaluation strictly follows a cascaded paradigm, where a failure at any stage terminates the episode immediately.
Let $N$ denote the total number of evaluation episodes. The evaluation formulation is defined as follows:
\begin{itemize}
    \item \textbf{Task Planning Success Rate ($SR_{\text{plan}}$):} 
    The execution sequence initiates with the task planning phase ($k=0$). The planning success rate is defined as:
    \begin{equation}
        SR_{\text{plan}} = \frac{N_{\text{plan}}}{N},
        \label{eq:psr_plan}
    \end{equation}
    where $N_{\text{plan}}$ is the number of episodes in which the task planner correctly identifies the requisite skill chain.
    \item \textbf{Success Rate of Individual Skill Completion:}
    To isolate the policy's inherent capability for each skill without the influence of cascading errors, we define the success rate of individual skill completion. This evaluates the execution of skill $S_k$ in isolation, decoupled from the task planner and preceding skills. Let $N_{S_k}$ denote the total number of independent trials initialized specifically for skill $S_k$, and $N_{S_k}^{\text{success}}$ denote the number of successful executions. The formulation is:
    \begin{equation}
        \text{Individual} = \frac{N_{S_k}^{\text{success}}}{N_{S_k}}.
        \label{eq:isr_k}
    \end{equation}
    This metric provides a fundamental evaluation for different policy models.
    % \item \textbf{Success Rate of Conditional Skill Completion:} 
    % For the subsequent skill execution stages ($k \in \{1, \dots, K\}$), the policy is evaluated conditionally. Let $P(S_k \mid S_{1:k-1}, \text{plan})$ denote the conditional probability of successfully completing the $k$-th skill, given that the initial planning and all preceding skills $S_{1:k-1}$ have been executed without error.
    
    \item \textbf{Progressive Skill Completion Success Rate:} 
    The Progressive Skill Completion Success Rate at skill $S_k$ is the fraction of episodes that successfully complete $S_k$ among those that have successfully completed all preceding skills.
    It effectively captures the cascading errors inherent in long-horizon tasks and is formally defined as the joint probability across the chain of skills:
    \begin{equation}
        PSCSR = SR_{\text{plan}} \prod_{i=1}^{K} P(S_i \mid S_{1:i-1}, \text{plan}).
        \label{eq:pcsr_k}
    \end{equation}
\end{itemize}

\begin{table*}[t]
  \centering
  \small
  \caption{\emphline{Numerical Results on the STMM-12k.} We report the success rate ($\%$) of task planning ($SR_{\text{plan}}$), individual skill completion ($\text{Individual}$), and progressive completion success rate of different skills across different settings. The higher the better. 
  \textit{(a)} \emphline{Evaluation of object relocation}. 
  \textit{(b)} \emphline{Evaluation of state transition}. 
  \textit{(c)} \emphline{Evaluation of multi-event setting}. Success rates are reported in the format ``First Sound Source / Second Sound Source''.
  $^*$ denotes the target position that is provided for the navigate policy model.
  }
  \label{tab:main_results}
    % \vspace{-.05in}
  % --- Row 1: (a) object relocation & (b) state transition ---
  
  % (a) object relocation (Left)
    \begin{subtable}[t]{0.47\linewidth}
        \centering
        \caption{}
        \resizebox{\linewidth}{!}{%
        \begin{tabular}{cccccc}
        \shline
        Method & $SR_{\text{plan}}$ & Navigate & Pick & Place & CSR \\
        \hline
        Individual & N/A & 81.02 & 96.59 & 91.97 & N/A \\
        Oracle & 100.00 & 81.02 & 95.50 & 92.77 & 71.78 \\
        Oracle$^*$ & 100.00 & 95.13 & 96.93 & 93.40 & 86.13 \\
        \hline
        \hline
        Qwen2.5-Omni-3B & 1.46 & 83.33 & 100.00 & 80.00 & 0.97 \\
        Qwen2.5-Omni-7B & 2.68 & 90.91 & 100.00 & 90.00 & 2.19 \\
        MiniCPM-o-4.5 & 13.38 & 80.00 & 97.73 & 88.37 & 9.25 \\
        Qwen3-Omni-30B & 88.56 & 81.32 & 95.27 & 92.55 & 63.50 \\
        SSP & 99.03 & 80.84 & 95.74 & 92.70 & \textbf{71.05} \\
        \hline
        \hline
        Qwen2.5-Omni-3B$^*$ & 1.46 & 83.33 & 100.00 & 100.00 & 1.22 \\
        Qwen2.5-Omni-7B$^*$ & 2.68 & 90.91 & 100.00 & 100.00 & 2.43 \\
        MiniCPM-o-4.5$^*$ & 13.38 & 96.36 & 96.23 & 90.20 & 11.19 \\
        Qwen3-Omni-30B$^*$ & 88.56 & 95.05 & 96.53 & 92.81 & 75.43 \\
        SSP$^*$ & 99.03 & 95.09 & 96.90 & 93.33 & \textbf{85.16} \\
        \shline
        \end{tabular}%
        }
        \label{tab:sota_stow}
    \end{subtable}%
        \hfill
    % (b) state transition (Right)
    \begin{subtable}[t]{0.47\linewidth}
        \centering
        \caption{}
        \resizebox{\linewidth}{!}{%
        \begin{tabular}{ccccc}
        \shline
        Method & $SR_{\text{plan}}$ & Navigate & Interaction & CSR \\
        \hline
        Individual & N/A & 44.20 & 94.20 & N/A \\
        Oracle & 100.00 & 44.20 & 65.25 & 28.84 \\
        Oracle$^*$ & 100.00 & 49.49 & 61.38 & 30.38 \\
        \hline
        \hline
        Qwen2.5-Omni-3B & 12.29 & 62.50 & 53.33 & 4.10 \\
        Qwen2.5-Omni-7B & 31.91 & 56.15 & 52.38 & 9.39 \\
        MiniCPM-o-4.5 & 41.47 & 48.97 & 59.66 & 12.12 \\
        Qwen3-Omni-30B & 39.42 & 48.05 & 64.86 & 12.29 \\
        SSP & 99.49 & 44.25 & 65.12 & \textbf{28.67} \\
        \hline
        \hline
        Qwen2.5-Omni-3B$^*$ & 12.29 & 69.44 & 46.00 & 3.92 \\
        Qwen2.5-Omni-7B$^*$ & 31.91 & 58.82 & 53.64 & 10.07 \\
        MiniCPM-o-4.5$^*$ & 41.47 & 53.50 & 58.46 & 12.97 \\
        Qwen3-Omni-30B$^*$ & 39.42 & 53.68 & 56.45 & 11.95 \\
        SSP$^*$ & 99.49 & 49.57 & 61.25 & \textbf{30.20} \\
        \shline
        \end{tabular}%
        }
        \label{tab:sota_interact}
    \end{subtable}%

  \vspace{1em}

  % --- Row 2: (c) multi-event setting (Center) ---
    \begin{subtable}[t]{0.90\linewidth}
        \centering
        \caption{}
        \resizebox{\linewidth}{!}{%
        \begin{tabular}{ccccccc}
            \shline
            Method & $SR_{\text{plan}}$ & Navigate & Pick & Place & Interaction & CSR \\
            \hline
            Individual & N/A & 73.75 / 64.50 & 65.82 / 44.14 & 54.75 / 38.28 & 89.29 / 84.72 & N/A \\
            Oracle & 100.00 & 69.25 / 39.16 & 73.58 / 57.50 & 75.69 / 91.30 & 93.55 / 96.00 & 41.50 / 11.25 \\
            Oracle$^*$ & 100.00 & 73.75 / 63.32 & 82.14 / 52.44 & 76.81 / 86.05 & 93.02 / 88.64 & 49.75 / 19.00 \\
            \hline
            \hline
            Qwen2.5-Omni-3B & 26.25 & 80.95 / 36.73 & 71.60 / 57.14 & 77.59 / 75.00 & 100.00 / 100.00 & 12.25 / 2.50 \\
            Qwen2.5-Omni-7B & 44.50 & 75.28 / 40.28 & 65.65 / 48.15 & 80.23 / 84.62 & 100.00 / 100.00 & 18.00 / 3.25 \\
            MiniCPM-o-4.5 & 50.00 & 71.00 / 40.96 & 72.73 / 58.62 & 76.04 / 88.24 & 100.00 / 100.00 & 20.75 / 5.00 \\
            Qwen3-Omni-30B & 19.25 & 62.34 / 50.00 & 60.00 / 75.00 & 87.50 / 100.00 & 87.50 / 100.00 & 7.00 / 3.00 \\
            SSP & 69.50 & 68.35 / 40.19 & 71.76 / 58.33 & 72.95 / 92.86 & 90.00 / 94.74 & \textbf{26.75} / \textbf{7.75} \\
            \hline
            \hline
            Qwen2.5-Omni-3B$^*$ & 26.25 & 79.05 / 53.70 & 80.77 / 41.67 & 77.78 / 60.00 & 100.00 / 80.00 & 13.50 / 2.50 \\
            Qwen2.5-Omni-7B$^*$ & 44.50 & 73.60 / 60.47 & 78.74 / 48.00 & 82.00 / 79.17 & 100.00 / 100.00 & 21.50 / 5.25 \\
            MiniCPM-o-4.5$^*$ & 50.00 & 74.50 / 68.32 & 81.95 / 50.00 & 77.98 / 86.21 & 100.00 / 90.91 & 25.25 / 8.75 \\
            Qwen3-Omni-30B$^*$ & 19.25 & 72.73 / 72.97 & 66.67 / 46.15 & 85.71 / 100.00 & 92.86 / 78.57 & 9.25 / 4.25 \\
            SSP$^*$ & 69.50 & 73.38 / 63.43 & 82.08 / 52.83 & 74.65 / 89.29 & 90.32 / 87.50 & \textbf{33.50} / \textbf{13.25} \\
            \shline
        \end{tabular}%
        }
        \label{tab:sota_two_sound}
    \end{subtable}%
  % \vspace{-.1in}
\end{table*}

\subsubsection{Result}
We provide detailed numerical results on the STMM-12k for quantitative comparison. In addition to the numbers shown in~\reffig{fig:sota} of the main paper, we add the success rate of individual skill completion (see Individual in~\reftab{tab:main_results}) and Progressive Skill Completion Success Rate for each skill.

\subsection{Details of low-level policy and Setting-Specific Planner}
\label{sec:policy_detail}
\subsubsection{Architecture}
\begin{figure}[t]
    \centering

    % 第一行
    \begin{subfigure}[b]{0.48\textwidth}
        \centering
        \includegraphics[width=\linewidth]{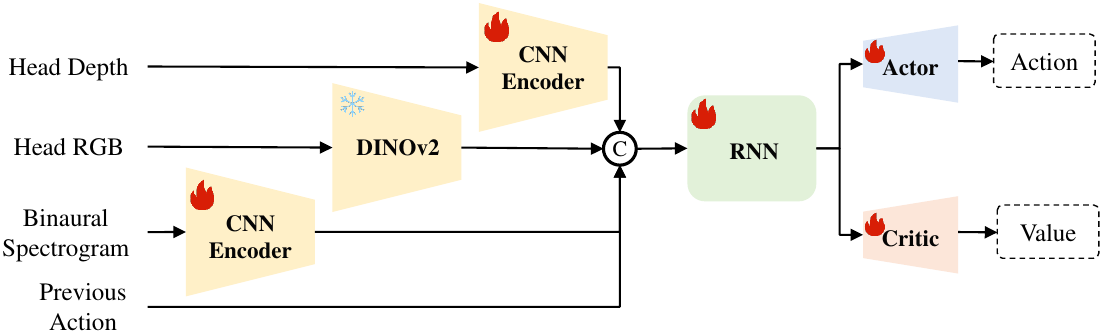}
        \caption{}
        \label{fig:supp_one_sound_nav_network}
    \end{subfigure}
    \hfill
    \begin{subfigure}[b]{0.48\textwidth}
        \centering
        \includegraphics[width=\linewidth]{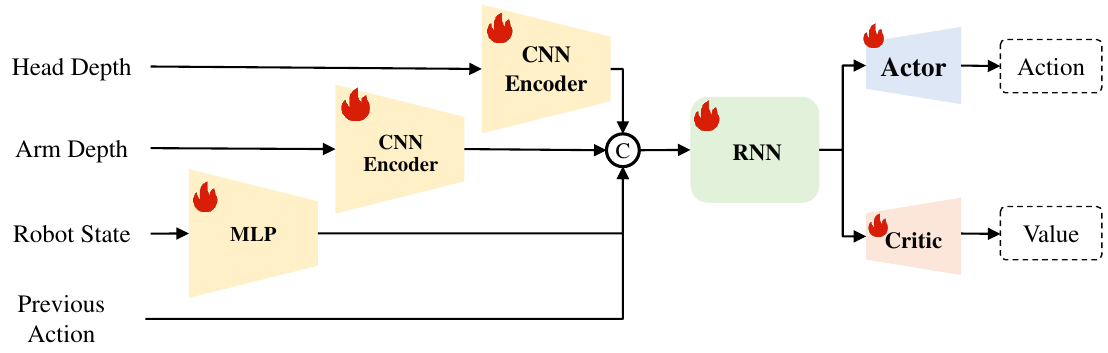}
        \caption{}
        \label{fig:supp_interact_network}
    \end{subfigure}

    \vspace{0.3cm}

    % 第二行
    \begin{subfigure}[b]{0.48\textwidth}
        \centering
        \includegraphics[width=\linewidth]{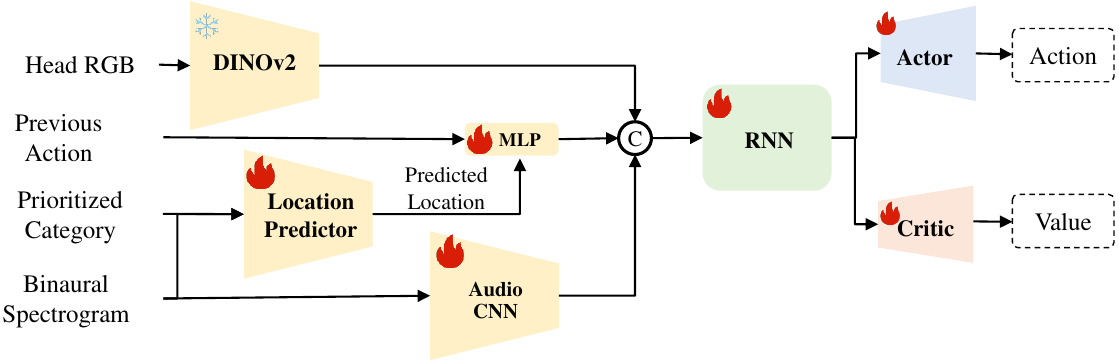}
        \caption{}
        \label{fig:supp_2sound_nav_network}
    \end{subfigure}
    \hfill
    \begin{subfigure}[b]{0.48\textwidth}
    \centering
    \raisebox{4ex}{%
        \includegraphics[
            width=0.6\linewidth,
            trim=0cm 0cm 0cm 0cm,
            clip
        ]{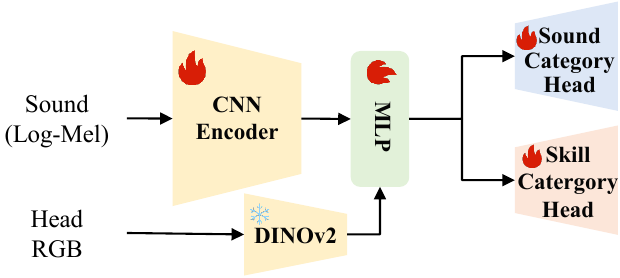}
    }
    \caption{}
    \label{fig:supp_one_sound_task_planner_network}
    \end{subfigure}

    \caption{
    \textbf{Architecture of low-level policies and Setting-Specific Planner (SSP)}.
    (a) Navigation policy at the single-event setting.
    (b) Low-level policy at the state transition task.
    (c) Navigation policy at the multi-event setting.
    (d) SSP at the single-event setting.
    }
    \label{fig:supp_network}
\end{figure}

We provide the architecture of the low-level policies and the Setting-Specific Planner (SSP) in~\reffig{fig:supp_network}.
For the navigation model in the single-event setting, multi-modal observations are processed via the corresponding encoder (see~\reffig{fig:supp_one_sound_nav_network}).
The policies for interaction skills, such as Open Door, Close Sink, and Open Washing Machine, share a unified network architecture (see~\reffig{fig:supp_interact_network}).
For the navigation model in the multi-event setting, an additional location predictor is introduced (see~\reffig{fig:supp_2sound_nav_network}).
All low-level policies are trained via PPO, with the internal location predictor (see~\reffig{fig:supp_2sound_nav_network}) trained concurrently via supervised learning.
The architecture of the Setting-Specific Planner (SSP) is shown in~\reffig{fig:supp_one_sound_task_planner_network}, which is trained in a supervised manner. 
The SSP processes single-channel mel spectrum and head RGB for the single-event setting, and priority-conditioned features for the multi-event setting.

\subsubsection{Reward and loss function}
All policies are trained via PPO. Reward functions are detailed below.

\noindent\textbf{Reward for Navigation Policy.}
We train the navigation policy using a dense progress reward, with additional penalties for time and collisions. At step $t$, the reward is defined as
\begin{equation}
\begin{aligned}
r_t ={}& -0.002
    + \left(d_{t-1}-d_t\right)
    - \min\left(0.001\Delta F_t,0.2\right)
    + 5.0\mathbb{I}_{\mathrm{succ},t},
  \label{eq:nav_reward}
  \end{aligned}
  \end{equation}
where $-0.002$ is a slack penalty applied at each step. 
$d_t$ denotes the shortest collision-safe geodesic distance from the current position to the nearest valid navigation goal at timestep $t$, such that progress toward the goal yields a positive reward and movement away from it incurs a penalty. 
$\Delta F_t = max((F_t-F_{t-1}), 0)$ denotes the increase in robot collision force, with the corresponding penalty capped at $0.2$ per step.
$\mathbb{I}_{\mathrm{succ},t}$ is the success indicator, which is activated only when the policy stops within the goal region.

\noindent\textbf{Reward for Low-level Policy.}
We train the low-level policy for the state-transition task using a phase-aware dense reward. The task consists of four phases: approaching, state transition, releasing, and returning the gripper to its resting pose. The reward at step $t$ is defined as
\begin{equation}
\begin{aligned}
r_t ={}&
-0.01
+ s_{\mathrm{approaching}} \left(d^{\mathrm{handle}}_{t-1}-d^{\mathrm{handle}}_t\right)
+ \mathbb{I}_{\mathrm{first\_grasp},t} \\
&+ s_{\mathrm{transiting}}\left(d^{\mathrm{joint}}_{t-1}-d^{\mathrm{joint}}_t\right)
+ 3\,\mathbb{I}_{\mathrm{first\_goal},t}
+ 2\,\mathbb{I}_{\mathrm{valid\_release},t} \\
&+ s_{\mathrm{reseting}}
    \operatorname{round}_{10^{-3}}\!
    \left(d^{\mathrm{rest}}_{t-1}-d^{\mathrm{rest}}_t\right) \\
&- \mathbb{I}_{\mathrm{wrong\_grasp},t}
- 2\,\mathbb{I}_{\mathrm{invalid\_release},t}
- 0.5\,\mathbb{I}_{\mathrm{invalid\_grasp},t} \\
&- 0.02\,\mathbb{I}\left(F_t \geq 1000\right)
+ 5\,\mathbb{I}_{\mathrm{success},t}.
\label{eq:dishwasher_reward}
\end{aligned}
\end{equation}

Here, 
$d^{\mathrm{handle}}_t$ denotes the distance from the gripper to the interactive region of the articulated object, 
$d^{\mathrm{joint}}_t = |\hat{q}-q_t|$ denotes the error between the target and current joint states, 
and $d^{\mathrm{rest}}_t$ denotes the distance from the gripper to its resting position. The phase-dependent factors $s_{\mathrm{approaching}}$, $s_{\mathrm{transiting}}$, and $s_{\mathrm{reseting}}$ scale the corresponding progress rewards.

The indicators $\mathbb{I}_{\mathrm{first\_grasp},t}$, $\mathbb{I}_{\mathrm{first\_goal},t}$, and $\mathbb{I}_{\mathrm{valid\_release},t}$ provide one-time bonuses for the first correct grasp, reaching the target state while maintaining the grasp, and releasing after reaching the target state, respectively. Wrong grasps, invalid releases, and invalid grasps incur penalties and terminate the episode. A collision force exceeding $1000$N incurs an additional penalty. Success requires the object to reach its target state, the gripper to release it, and the gripper to return within $0.05$m of its resting position.

\noindent\textbf{PPO Loss.}
Given the task-specific reward $r_t$, we compute
the return target $\hat{R}_t$ using Generalized Advantage Estimation (GAE) and
define the advantage as
\begin{equation}
\hat{A}_t = \hat{R}_t - V(x_t),
\end{equation}
where $x_t$ is the policy input and $V(x_t)$ is the critic prediction.

The PPO loss is defined as
\begin{equation}
\begin{aligned}
\mathcal{L}_{\mathrm{PPO}}
={}&
-\mathbb{E}_t\left[
\min\left(
\rho_t\hat{A}_t,\,
\operatorname{clip}(\rho_t,1-\epsilon,1+\epsilon)\hat{A}_t
\right)
\right] \\
&+ c_v\,\mathbb{E}_t\left[
\left(V(x_t)-\hat{R}_t\right)^2
\right]
-\beta\,\mathbb{E}_t\left[
\mathcal{H}\bigl(\pi(\cdot\mid x_t)\bigr)
\right],
\label{eq:ppo_loss}
\end{aligned}
\end{equation}
where
\begin{equation}
\rho_t =
\frac{\pi(a_t\mid x_t)}
{\pi_{{\mathrm{old}}}(a_t\mid x_t)}.
\end{equation}
Here, the first term updates the action policy according to the advantage,
the second term trains the critic to predict the reward-induced return, and
the entropy term $\mathcal{H}$ encourages exploration. Different policies share the same optimization objective, while their different reward functions lead to different return targets and
action updates.

\subsection{RIRs for Sound Rendering}
\label{sec:rir}
RIRs ensure two critical physical factors for sound rendering:
(1) {Geometry-aware propagation:} Through bidirectional path tracing, RIRs capture intricate wave phenomena, including direct paths, specular reflections, and reverberation. This ensures that the rendered sounds remain consistent with both the room's static 3D geometry and the dynamic position and orientation of the robot.
(2) {Material-dependent scattering:} RIRs account for acoustic material properties by mapping semantic annotations to frequency-dependent absorption and transmission coefficients.
This enables the simulator to reflect distinct acoustic interactions with diverse surfaces.

\subsection{Limitation}
\label{sec:limitation}
This initial exploration also highlights several limitations and directions for future work. Our evaluation focuses on simulation, leaving real-robot validation as an important next step. Additionally, state transitions remain challenging, as do task planning and navigation to the second sound source in multi-event tasks. Incorporating recovery behaviors into the navigation skill may help the robot recover when stuck by obstacles and continue navigating toward the second sound source.

\end{document}